\documentclass{article}

\usepackage[main, final]{neurips_2026}

\usepackage[utf8]{inputenc} % allow utf-8 input
\usepackage[T1]{fontenc}    % use 8-bit T1 fonts
\usepackage{hyperref}       % hyperlinks
\usepackage{url}            % simple URL typesetting
\usepackage{booktabs}       % professional-quality tables
\usepackage{amsfonts}       % blackboard math symbols
\usepackage{nicefrac}       % compact symbols for 1/2, etc.
\usepackage{microtype}      % microtypography
\usepackage{xcolor}         % colors
\usepackage{bbm}
\usepackage{bm}
\usepackage{multirow}
\usepackage{tabularx}
\usepackage{longtable}
\usepackage[table]{xcolor}
\usepackage{graphicx}
\usepackage{amsmath}
\usepackage{mathtools}
\usepackage{float}
\usepackage{titlesec}
\usepackage{subcaption}
\usepackage[shortlabels]{enumitem}
\usepackage{bbm}
\usepackage{amsthm}
\theoremstyle{definition}
\newtheorem{definition}{Definition}
\newtheorem{assumption}{Assumption}
\newtheorem{lemma}{Lemma}
\newtheorem{theorem}{Theorem}
\newtheorem{corollary}{Corollary}
\newtheorem{proposition}{Proposition}
\newtheorem{remark}{Remark}

\title{Accuracy of Noise Ceiling Estimators for Brain Scores}

\author{%
    Aude Maier \And Lucas Gruaz \And Martin Schrimpf\thanks{Joint supervision} \And Johanni Brea\footnotemark[1]\AND
  EPFL, Lausanne, Switzerland\vspace{5pt}\\
\texttt{\{aude.maier,lucas.gruaz,martin.schrimpf,johanni.brea\}@epfl.ch}
}

\titlespacing*{\section}      {0pt}{4pt}{1pt}
\titlespacing*{\subsection}   {0pt}{4pt}{1pt}
\titlespacing*{\subsubsection}{0pt}{0pt}{0pt}
\titlespacing{\paragraph}{%
  0pt}{%              left margin
  0.5\baselineskip}{% space before (vertical)
  1em}%               space after (horizontal)

\begin{document}

\maketitle

\begin{abstract}
    The internal activations of modern machine learning models align surprisingly well with the activities recorded in real brains. In the last five years, the alignment between encoding models and brain activities, often called brain scores, has steadily increased. Because brain recordings are noisy, there exists an upper bound to the brain scores that any model can achieve. This bound, called the noise ceiling, is crucial for two reasons: it determines whether further improvements in model performance are possible, and it enables meaningful comparisons of brain scores across benchmarks. The field has developed multiple approaches to estimate noise ceilings, but the statistical properties of existing estimators remain poorly understood, leaving it unclear in which experimental settings they provide reliable estimates. Here, we introduce a unified mathematical framework for brain recordings, use it to derive closed-form expressions for four commonly used noise ceiling estimators, and provide theoretical predictions for the behavior of each estimator.
    We validate the theoretical predictions with simulations that we compare to noise ceiling estimates in existing fMRI and EEG datasets.
    We find that the existing methods underestimate the true noise ceiling in different settings.
    We conclude with recommendations for best practice in different situations.
\end{abstract}

\section{Introduction}\label{sec:introduction}
A central goal of NeuroAI is to build computational models that not only solve tasks the brain solves, but do so with internal representations that match those of biological neural circuits \citep{yaminsUsingGoaldrivenDeep2016,schrimpfIntegrativeBenchmarkingAdvance2020,doerigNeuroconnectionistResearchProgramme2023,sadehEmergenceNeuroAIBridging2025}. The dominant empirical instrument for testing this match is the brain score: a measure of how well a model's internal activations predict neural responses to a shared stimulus set, typically via a linear readout fit on a training split and evaluated on held-out data \citep[e.g.,][]{schrimpfBrainScoreWhichArtificial2020}. Since the first demonstrations that task-optimized CNNs predict ventral-stream responses better than hand-designed models, brain scores have proliferated across modalities: vision \citep{yaminsPerformanceoptimizedHierarchicalModels2014,khaligh-razaviDeepSupervisedNot2014,schrimpfBrainScoreWhichArtificial2020,cadenaDeepConvolutionalModels2019}, audition \citep{kellTaskOptimizedNeuralNetwork2018,tuckuteManyNotAll2023}, and language \citep{wehbeLanguageProcessingBrain2019,schrimpfNeuralArchitectureLanguage2021,caucheteuxEvidencePredictiveCoding2023,leiLargeLanguageModels2026,alkhamissiLanguageCognitionHow2025}. They have substantially improved as architectures and training data have scaled \citep{antonelloScalingLawsLanguage2023,gokceScalingLawsTaskOptimized2025}.

Brain scores should be measured relative to what is achievable. Neural recordings are noisy: trial-to-trial variability, measurement noise, and finite sampling all place a hard ceiling on the variance any deterministic model can hope to predict \citep{hsuQuantifyingVariabilityNeural2004, sahaniHowLinearAre2002,schoppeMeasuringPerformanceNeural2016}. This noise ceiling is what separates a model that has captured the stimulus-driven signal from one that has room to improve. Without it, absolute brain scores are hard to interpret, as they confound model quality with data reliability. The noise ceiling reframes the question from ``how high is the score?'' to ``how much of the explainable variance has been explained?'', a quantity on which models can be fairly compared across datasets, regions, and recording modalities.

Estimating the noise ceiling, however, is harder than it looks. The literature contains multiple procedures. \citet{sahaniHowLinearAre2002} introduced the explainable signal power estimator (SPE) to measure the fraction of stimulus-related variance of neural responses captured by models. \citet{hsuQuantifyingVariabilityNeural2004} introduced the normalized correlation coefficient $CC_\text{norm}$, which expresses model performance as a fraction of an estimated upper bound $CC_\text{max}$ computed via split-half reliability with Spearman–Brown correction. \citet{schoppeMeasuringPerformanceNeural2016} showed that  $CC_\text{norm}$ is preferable over the SPE in certain settings, and derived an alternative way of computing $CC_\text{max}$ from signal power without resampling, avoiding computational expenses of the split-half procedure.
\citet{lage-castellanosMethodsComputingMaximum2019} extended this analysis to fMRI, comparing Monte Carlo, split-half, and analytic noise ceiling estimators across noise structures and regularization regimes.
\citet{schrimpfNeuralArchitectureLanguage2021} introduced an extrapolation procedure used in Brain-Score. \citet{vanbreeHowMuchVariance2025} clarified a common misinterpretation of noise ceilings in model evaluation, showing that split-half reliability is often incorrectly treated as a maximum achievable correlation rather than a bound on explainable variance ($R^2$). They demonstrate that this confusion can lead to inflated or misleading scores for model performance.

In this work, we revisit the noise ceiling from the ground up. We review the main families of estimators, derive analytical expressions under a generative model of brain signals, and characterize the assumptions and failure modes of each. Empirically, we show the qualitative match between model predictions and noise ceiling estimates on fMRI and EEG data. We close with concrete recommendations for which estimator to use, with the aim of putting brain-score comparisons on firmer statistical footing.

\section{Background}\label{sec:background}

Brain recordings obtained from a stimulus-driven experiment can be represented as a set of vectors $\{\mathbf{v}_1,\ldots,\mathbf{v}_m\},\,\mathbf{v}_i\in\mathbb{R}^T$. Each vector $\mathbf{v}_i$ corresponds to the response of a single recording unit, e.g. a voxel in the case of fMRI or an electrode in the case of EEG. If the experiment involves a single continuous stimulus (e.g., watching a movie or listening to a story), $T$ is the duration of the stimulus divided by the temporal resolution of the recording device, and $v_{i,t}$  denotes the signal of recording unit $i$ at timestep $t$. For experiments consisting of a sequence of discrete stimuli, each index $t$ designates one stimulus and $T$ is the total number of stimuli. Noise ceiling computations rely on experiments where the same stimuli are presented multiple times, and the resulting trials are used to assess the consistency of the brain response. We denote by $\mathbf{v}_i^{(\ell)},\,\ell\in\{1,\ldots,n\}$ the signal recorded from unit $i$ during trial $\ell$. Depending on the experimental setting, the trials can refer to multiple presentations of the same stimuli to the same subject, or to single presentations of the stimuli to different subjects. While these two settings have different statistical interpretations, which we discuss later, they can be handled within a unified framework.

\paragraph{Encoding Models and Brain Scores} Encoding models aim to provide a mapping from the stimulus space to the brain recording space, i.e. predicting the brain response $\hat{\mathbf{v}}_i$ for each recording unit $i$ given the sequence of stimuli $t=1,\ldots,T$. In this work, we focus on encoding models and do not consider the Representational Similarity Analysis (RSA) framework. The performance of such a model is usually evaluated in a two-step procedure: i) compute per unit $i$ a similarity $S_i$ between the predictions $\hat{\mathbf{v}}_i$ and the actual recordings $\mathbf{v}_i$, and ii) compute the average score $S = \frac1m\sum_{i=1}^m \frac{S_i}{\widehat{\mathrm{NC}}_i}$, relative to some estimate $\widehat{\mathrm{NC}}_i$ of the noise ceiling of unit $i$. The most common similarity measure is the Pearson correlation\footnote{$\widehat{\mathrm{Corr}}_t(\mathbf{x}, \mathbf{y}) = \frac1T\sum_{t=1}^T(x_t - \mu_x)(y_t - \mu_ y)/(\sigma_x\sigma_y)$, with average $\mu_x$ and empirical variance $\sigma_x^2$ computed over time} $S_i = \widehat{\mathrm{Corr}}_t(\hat{\mathbf{v}}_i, \mathbf{v}_i)$.
The correlation can be computed either at the level of individual trials, $\widehat{\mathrm{Corr}}_t\left(\hat{\mathbf{v}}_i^{(\ell)}, \mathbf{v}_i^{(\ell)}\right)$, or on the signal averaged across trials $\widehat{\mathrm{Corr}}_t\left(\hat{\mathbf{v}}_i, \overline{\mathbf{v}}_i\right)$. The choice between these two approaches is usually driven by the experimental setting: within-subject repetitions are typically averaged to reduce noise, while across-subject trials are more commonly kept separate, both to preserve idiosyncratic signal and due to the difficulty of establishing a meaningful correspondence between recording units across subjects. Encoding models can learn a separate mapping for each subject through a training procedure on a held-out portion of the recorded signal. However, if the recordings contain variability that is neither deterministically driven by the stimulus nor captured by trial-specific parameters, no encoding model can achieve a perfect score.
\par
\paragraph{Noise Ceiling} If we had access to the noise-free signal $\mathbf{v}_i^{*(\ell)}$ we could compute the noise ceiling for recording unit $i$ as
\begin{equation}
    NC_i^{\mathrm{mean}}=\widehat{\mathrm{Corr}}_t\left(\overline{\mathbf{v}}^*_i, \overline{\mathbf{v}}_i\right)\, ,\qquad NC_{i,\ell}^{\mathrm{single}}=\widehat{\mathrm{Corr}}_t\left({\mathbf{v}}_i^{*(\ell)}, \mathbf{v}_i^{(\ell)}\right)\, ,\label{eq:noise ceiling definition}
\end{equation}
depending on whether the encoding model predicts trial-averaged responses or individual trials. This noise ceiling would constitute an upper bound on the score achievable by any encoding model on a given dataset $\{\mathbf{v}_1^{(\ell)},\ldots,\mathbf{v}_m^{(\ell)}\}_{\ell=1}^n$. However, since the noise-free signals $\mathbf{v}_i^{*(\ell)}$ are unknown, the noise ceiling cannot be computed directly. Several methods have been proposed to estimate it from the observed recordings alone. Five such methods are commonly used in the literature and fall into two families: those targeting $NC^{\mathrm{mean}}$ and those targeting $NC^{\mathrm{single}}$. Table~\ref{table} provides a list of papers that use each of these methods. We describe each of these methods in the remainder of this section.

\subsection{Noise Ceiling for Trial-Averaged Responses}

Two methods are commonly used to estimate $NC^{\mathrm{mean}}$. Both assume each recording decomposes as $\mathbf{v}_i^{(\ell)}=\mathbf{v}_i^{*}+\boldsymbol{\eta}_i^{(\ell)}$, where $\mathbf{v}_i^{*}$ is perfectly shared across trials and $\boldsymbol{\eta}_i^{(\ell)}$ are i.i.d.\ with variance $\sigma^2_\eta$ \citep{hsuQuantifyingVariabilityNeural2004, schoppeMeasuringPerformanceNeural2016}. Under this model and in the limit of long recordings, the true noise ceiling is
\begin{equation}
    NC_i^{\mathrm{mean}}=\widehat{\mathrm{Corr}}_t\left(\mathbf{v}^*_i, \mathbf{v}^*_i+\overline{\boldsymbol{\eta}}_i\right)=\sqrt{\frac{\sigma_s^2}{\sigma_s^2+\frac{\sigma_\eta^2}{n}}}\label{eq:nc_simple}\, ,
\end{equation}
where $\sigma_s^2 = \widehat{\mathrm{Var}}_t(\mathbf{v}_i^*)$.
In the \textbf{Split-Half Reliability with Spearman-Brown Correction (SHSB)} method,
the $n$ trials are split into two equal halves and their respective averages correlated. Since each half-average contains twice the noise of the full average, the Spearman-Brown correction is applied \citep{hsuQuantifyingVariabilityNeural2004,vanbreeHowMuchVariance2025}. The procedure is repeated over multiple random splits and results are averaged (\autoref{tab:nc_methods}).
An alternative approach \citep{schoppeMeasuringPerformanceNeural2016,lage-castellanosMethodsComputingMaximum2019,allenMassive7TFMRI2022}, the \textbf{Across-Trial Variance method (ATV)}, estimates the signal variance using $\sigma_s^2 = \sigma_{\mathrm{total}}^2 - \sigma_\eta^2/n$, where the total variance $\sigma_{\mathrm{total}}^2$ is estimated as $\widehat{\mathrm{Var}}_t(\overline{\mathbf{v}}_i)$ and the noise variance $\sigma_\eta^2$ as the average across-trial variance at each timestep (\autoref{tab:nc_methods}).

\subsection{Noise Ceiling for Single-Trial Responses}

Some experiments involve a single long continuous recording per subject (e.g.\ a movie or story \citep{freteaultAlignmentAuditoryArtificial2025, chenSharedMemoriesReveal2017, wehbeLanguageProcessingBrain2019}), with trials corresponding to different subjects. For low-variability sensory areas, researchers average across subjects and apply SHSB or ATV \citep{tangDiversePerceptualRepresentations2025, tuckuteDrivingSuppressingHuman2024}. For higher-level areas (e.g.\ language), inter-subject variability is substantial and anatomical misalignment may cancel shared signals, so encoding models are fit per subject and $NC^{\mathrm{single}}$ must be estimated instead. We summarize three commonly used estimators.
\textbf{Leave-One-Out Correlation (LOOC)}:
each trial $\mathbf{v}_i^{(\ell)}$ is correlated against the leave-one-out average $\overline{\mathbf{v}}_i^{(-\ell)}\coloneq\frac{1}{n-1}\sum_{\ell'\neq\ell}\mathbf{v}_i^{(\ell')}$, and results are averaged over trials.
\textbf{Trial-To-Mean Correlation (TTMC)}:
same as LOOC, except the individual trial is included in the average it is compared against.
\textbf{Leave-One-Out Prediction (LOOP)}
introduced by \cite{schrimpfNeuralArchitectureLanguage2021}, a linear model is trained on a pool of $n$ trials to predict the held-out trial $\mathbf{v}_i^{(\ell)}$, mirroring the encoding model training procedure. The pool size $n$ is varied, and the noise ceiling extrapolated to $n\rightarrow\infty$ by fitting $\widehat{NC}_{i,\ell}^{\mathrm{LOOP}}(n)=\widehat{NC}_{i,\ell}^{\infty}\!\left(1-e^{-n/\tau_0}\right)$ (\autoref{tab:nc_methods}).

\begin{table}[htbp]
\centering
\caption{Noise ceiling estimator formulas. Negative values under the square root are thresholded at zero; SHSB additionally averages over multiple random splits. $\overline{\mathbf{v}}$: trial average, $\overline{\mathbf{v}}_i^{(-\ell)}$: trial average excluding trial $\ell$, $\overline{\mathbf{v}}_i^{(1)}$: average over (randomly selected) first half of trials, $\widehat{\mathbf{v}}^{(\ell)}_i$: linear prediction given all other trials}
\label{tab:nc_methods}
\renewcommand{\arraystretch}{2}\small
\scalebox{1}{
\begin{tabular}{ll||ll}
 & \textbf{Trial-averaged} $\left(\widehat{NC}^{\mathrm{mean}}_i\right)$ &
 & \textbf{Single-trial} $\left(\widehat{NC}^{\mathrm{single}}_{i,\ell}\right)$ \\
\midrule
SHSB &
  $\sqrt{\dfrac{2\,\widehat{\mathrm{Corr}}_t\!\left(\overline{\mathbf{v}}_i^{(1)},\overline{\mathbf{v}}_i^{(2)}\right)}{1+\widehat{\mathrm{Corr}}_t\!\left(\overline{\mathbf{v}}_i^{(1)},\overline{\mathbf{v}}_i^{(2)}\right)}}$\quad
  & LOOC &
  $\widehat{\mathrm{Corr}}_t\!\left(\mathbf{v}_i^{(\ell)},\overline{\mathbf{v}}_i^{(-\ell)}\right)$  \\
  &
     & TTMC & $\widehat{\mathrm{Corr}}_t\!\left(\mathbf{v}_i^{(\ell)},\overline{\mathbf{v}}_i\right)$\\[0.5ex]
     ATV  &
  $\sqrt{\dfrac{\widehat{\mathrm{Var}}_t\!\left(\overline{\mathbf{v}}_i\right)
    -\frac{1}{n}\frac{1}{T}\sum_t\widehat{\mathrm{Var}}_\ell\!\left(v_{i,t}^{(\ell)}\right)}
   {\widehat{\mathrm{Var}}_t\!\left(\overline{\mathbf{v}}_i\right)}}$ & LOOP &
  $\widehat{\mathrm{Corr}}_t\!\left(\mathbf{v}_i^{(\ell)},\widehat{\mathbf{v}}^{(\ell)}_i\right)$ \\

\end{tabular}}
 \vspace{-4pt}
\end{table}

\section{Results}\label{sec:results}
\subsection{Generative Model of Brain Recordings\label{sec:framework}}
We define a recording $\mathbf{v}^{(\ell)}_i = \{v^{(\ell)}_{i,t}\}_{t>0}$ as a stochastic process. Each trial $\ell$ carries attributes $\theta^{(\ell)}$ that are fixed within the trial but may differ between trials, such as the experimental setup, the internal state of the subject, or the identity of the subject.
We define the reproducible part of the signal as the conditional expectation $v^{*(\ell)}_{i,t} = \mathbb{E}\left[v_{i,t} \mid \theta^{(\ell)}\right]$, the expectation being taken over everything that is not reproducible within the trial, and the noise as $\eta^{(\ell)}_{i,t} = v^{(\ell)}_{i,t} - v^{*(\ell)}_{i,t}$. The attributes $\theta^{(\ell)}$ are identical for all $\ell$ only when the trials are repeated presentations under identical conditions, in which case $v^{*}_{i,t}$ is the same for every trial.
\par
The centered reproducible responses $\mathbf{v}_i^{*(\ell)}-\langle\mathbf{v}_i^{*(\ell)}\rangle$, where $\langle\cdot\rangle$ denotes the average over time, span a subspace of $\mathbb{R}^T$ of dimension $K\leq nm$. For any basis $\{\boldsymbol{\phi}_k\}_{k=1}^K$ of this subspace that is orthonormal over time, i.e. $\langle\boldsymbol{\phi}_k\boldsymbol{\phi}_{k'}\rangle=1$ if $k=k'$ and $0$ otherwise, the signal recorded from unit $i$ during trial $\ell$ can be written as
\begin{equation}
    \mathbf{v}_i^{(\ell)} = b_i^{(\ell)}+\sum_{k=1}^K w_{ik}^{(\ell)} \boldsymbol{\phi}_k + \boldsymbol{\eta}_i^{(\ell)},
    \qquad \mathbf{v}_i^{*(\ell)} = b_i^{(\ell)}+\sum_{k=1}^K w_{ik}^{(\ell)} \boldsymbol{\phi}_k,
    \label{eq: v}
\end{equation}
where $b_i^{(\ell)}:=\langle\mathbf{v}_i^{*(\ell)}\rangle$ is a constant offset, $w_{ik}^{(\ell)} := \widehat{\mathrm{Cov}}_t(\mathbf{v}_i^{*(\ell)}, \boldsymbol{\phi}_k)$ are trial- and unit-specific mixing weights, and $\boldsymbol{\eta}_i^{(\ell)}$ is the noise defined above.
\paragraph{Latent signals} This linear decomposition provides a useful interpretation: the basis signals $\boldsymbol{\phi}_k$ can be seen as latent signals of neural activity elicited by the stimulus. This interpretation builds on the view that stimulus-driven brain activity lives in a low-dimensional subspace of the full space of neural activity \citep{cunninghamDimensionalityReductionLargescale2014, gaoTheoryMultineuronalDimensionality2017}, so that $K$ is typically much smaller than the number of neurons, although it might exceed the number $m$ of measured units. Moreover, the inter-subject correlation literature shows that stimulus-driven responses are largely shared across subjects \citep{hassonIntersubjectSynchronizationCortical2004, simonyDynamicReconfigurationDefault2016}, suggesting that there exists a subspace of neural activity common to all subjects. This formulation is similar to the Shared Response Model \citep{chenReducedDimensionFMRIShared2015} and related functional alignment approaches \citep{haxbyHyperalignmentModelingShared2020}, which decompose the stimulus-driven component of fMRI recordings into shared latent signals and subject-specific mixing weights. It is also what encoding models presuppose: a single model produces features shared across subjects, and a subject-specific linear readout is fitted to each, implying that the reproducible responses lie in a common subspace spanned by the model features. In this view, $K$ is a lower bound on the number of features a model needs. The framework is intentionally kept general across recording modalities.

\paragraph{Mixing weights and regimes} The weights $w_{ik}^{(\ell)}$ describe how strongly each latent signal $\boldsymbol{\phi}_k$ is expressed in the reproducible response of unit $i$ during trial $\ell$. In the ideal case where every trial presents the same stimuli to the same subject under exactly the same conditions (i.e. the attributes $\theta^{(\ell)}$ are constant), the weights are identical for every trial $w_{ik}^{(\ell)}=w_{ik}$. We call this the \emph{homogeneous regime}. When trials correspond to different subjects, differences in anatomy and functional organization lead each subject to express the latent signals differently. Even within a single subject, changes in the experimental setting or in the internal state of the subject can change how the latent signals are expressed from one trial to the next. In both cases, the weights differ between trials, and we speak of the \emph{heterogeneous regime}. The homogeneous regime should therefore be seen as an idealization. Repeated trials on a single subject can approach it, whereas trials drawn from different subjects generally do not satisfy it.
\paragraph{Noise} The noise term $\boldsymbol{\eta}_i^{(\ell)}$ captures all variability in the recorded signal that is not reproducible within the trial. It can arise from several sources, such as the intrinsic stochasticity of neural processes, ongoing activity that is not driven by the stimulus, or the noise introduced by the recording device. It can exhibit temporal correlations, and its variance may depend on the signal and vary over time or between trials.
\par
By substituting the signal model and the noise-free component of \eqref{eq: v} into the noise ceiling estimators presented in Sec.~\ref{sec:background}, we can derive closed-form expressions for them. This allows us to assess, for each method, under what conditions it provides an accurate estimate of the true noise ceiling and where it may fall short. We carry out this analysis in the following paragraphs, treating the $NC^{\mathrm{mean}}$ and $NC^{\mathrm{single}}$ estimators in turn.
\paragraph{Notation and assumptions} For each unit $i$, we write $\bar{w}_{ik}:=\frac{1}{n}\sum_{\ell=1}^nw_{ik}^{(\ell)}$ for the average weights across trials, and $\widehat{\mathrm{Var}}_\ell\left[w_{ik}\right]:=\frac{1}{n}\sum_{\ell=1}^n\left(w_{ik}^{(\ell)}-\bar{w}_{ik}\right)^2$ for their variance. Similarly, we write $\sigma_{i,\ell}^2:=\frac{1}{T}\sum_{t=1}^T\mathrm{Var}\left(\eta_{i,t}^{(\ell)}\right)$ for the noise variance of trial $\ell$ averaged over time, and $\bar{\sigma}_i^2:=\frac{1}{n}\sum_{\ell=1}^n\sigma_{i,\ell}^2$ for its average over trials. All analytical results rely on four assumptions, stated formally in Appendix~\ref{sec:large_T_convergence}: (1) the noise of different trials $\boldsymbol{\eta}^{(\ell)}_i$, $\boldsymbol{\eta}^{(\ell')}_i$ is independent, conditionally on the attributes $\{\theta^{(\ell)}\}_{\ell=1}^n$ (the noise of different units within a trial may be correlated); (2) the noise-free signals $v_{i,t}^{*(\ell)}$ are bounded; (3) the noise $\eta^{(\ell)}_{i,t}$ has absolutely summable temporal autocovariance, i.e. its autocorrelation over time decays quickly; and (4) the squared noise ${\eta^{(\ell)}_{i,t}}^2$ has finite variance and absolutely summable temporal autocovariance. These assumptions are discussed in Sec.~\ref{sec:discussion}. Furthermore, the closed-form expressions below only hold for units $i$ whose stimulus-driven part is non-zero, i.e. $\widehat{\mathrm{Var}}_t\left(\bar{\mathbf{v}}^*_i\right)>0$ for trial-averaged noise ceiling and $\widehat{\mathrm{Var}}_t\left({\mathbf{v}}^{*(\ell)}_{i}\right)>0$ for single-trial noise ceiling, so that the quantities from \eqref{eq:noise ceiling definition} are well-defined.
\subsection{Noise Ceiling for Trial-Averaged Responses}\label{sec:results_th_mean}
As derived in Appendix~\ref{sec:true_noise_ceiling}, the true noise ceiling for trial-averaged responses of unit $i$ is
\begin{equation}
    NC_i^{\mathrm{mean}}=\sqrt{\frac{\sum_{k=1}^K\bar{w}_{ik}^2}{\sum_{k=1}^K\bar{w}_{ik}^2+\bar{\sigma}_i^2/n}}+\varepsilon_T                             ,\label{eq: nc_mean_main}
\end{equation}
where $\varepsilon_T$ is a random error term with bias $\left|\mathbb{E}\left[\varepsilon_T\right]\right|\leq c/T$ and mean squared error $\mathbb{E}\left[\varepsilon_T^2\right]\leq c/T$, for a constant $c$ that depends only on the bounds in our assumptions. For long recordings,  $NC_i^{\mathrm{mean}}$ approaches the leading term in \eqref{eq: nc_mean_main}, its random fluctuations around it shrink like $T^{-1/2}$, and its systematic bias like $T^{-1}$. One observes that the noise ceiling converges to $1$ as $n\rightarrow\infty$. Note that in the homogeneous regime, $\mathrm{\widehat{Var}}_\ell[w_{ik}]=0$ and \eqref{eq: nc_mean_main} converges to \eqref{eq:nc_simple}.
\paragraph{ATV estimator} As derived in Appendix~\ref{sec:ATV}, the ATV estimator is given by
\begingroup\makeatletter\def\f@size{8}\check@mathfonts
\begin{equation}
    \widehat{NC}_i^{\mathrm{ATV}}=\sqrt{\left[\frac{\sum_k\bar{w}_{ik}^2-\frac{1}{n-1}\sum_k\widehat{\mathrm{Var}}_\ell\left[w_{ik}\right]}{\sum_k\bar{w}_{ik}^2+\bar{\sigma}^2/n}\right]_+}+\varepsilon_T^{\mathrm{max}}=\sqrt{\left[\left(NC^{\mathrm{mean}}_i\right)^2-\frac{\frac{1}{n-1}\sum_k\widehat{\mathrm{Var}}_\ell\left[w_{ik}\right]}{\sum_k\bar{w}_{ik}^2+\bar{\sigma}^2/n}\right]_+}+\varepsilon_T^{\mathrm{max}},\label{eq:ATV main text}
\end{equation}
\endgroup
where $[x]_+:=\mathrm{max}(0,x)$ and the random error term satisfies $\left|\mathbb{E}\left[\varepsilon_T^{\mathrm{max}}\right]\right|\leq(c/T)^{1/4}$ and $\mathbb{E}\left[\left(\varepsilon_T^{\mathrm{max}}\right)^2\right]\leq(c/T)^{1/2}$, for a constant $c$ that depends neither on $T$ nor on $n$. These bounds are weaker than those of $\varepsilon_T$ because the threshold $\mathrm{max}(0,\cdot)$ is not differentiable at zero. Denoting by $x_i$ the argument of the square root, if $x_i$ is bounded away from zero uniformly in $T$, the error term becomes $\varepsilon_T$ and the $1/T$ rate is recovered. However, this rate is only reached once the fluctuations of the estimation of $x_i$ are small compared to $x_i$. For units with weak signal (small $x_i$), at small $T$ these fluctuations are large compared with the distance between $x_i$ and zero and the error is of order $T^{-1/4}$. In this transient regime, the error is dominated by an upward bias since the threshold sets to zero the realizations that are randomly too low, but keeps those that are randomly too high. Additionally, the leading term of $\widehat{NC}^{\mathrm{ATV}}_i$ underestimates the true noise ceiling by a term of order $\mathcal{O}\left(1/n\right)$, which vanishes in the homogeneous regime.
\paragraph{SHSB estimator}
We write $(A,B)$ for a split of the $n$ trials into two halves of size $n/2$, and $\mathcal{S}$ for the collection of $r$ splits over which the Spearman--Brown corrected correlations are averaged. As derived in Appendix~\ref{SHSB}, if for every split $\widehat{\mathrm{Cov}}_t\left(\mathbf{v}_i^{*A},\mathbf{v}_i^{*B}\right)\geq0$, then
\begingroup\makeatletter\def\f@size{10}\check@mathfonts
\begin{equation}
    \widehat{NC}_i^{\mathrm{SHSB}}=\sqrt{\frac{1}{r}\sum_{(A,B)\in\mathcal{S}}\frac{\sum_k\bar{w}_{ik}^2-\Delta_{AB}^{w}}{\sum_k\bar{w}_{ik}^2+\bar{\sigma}_i^2/n-\Delta_{AB}^{\mathrm{sd}}}}+\varepsilon_T^{\mathrm{max}},\label{eq:SHSB main text}
\end{equation}
\endgroup
where $\Delta_{AB}^{w}:=\frac{1}{4}\sum_k\left(\bar{w}_{ik}^A-\bar{w}_{ik}^B\right)^2$ and $\Delta_{AB}^{\mathrm{sd}}:=\frac{1}{4}\left(\sqrt{\sum_k\left(\bar{w}_{ik}^A\right)^2+\frac{2\bar{\sigma}_{A}^2}{n}}-\sqrt{\sum_k\left(\bar{w}_{ik}^B\right)^2+\frac{2\bar{\sigma}_{B}^2}{n}}\right)^2$ with $\bar{\sigma}_A^2:=\frac{2}{n}\sum_{\ell\in A}\sigma_{i,\ell}^2$ (and similarly for $B$). The term $\Delta_{AB}^{w}$ measures how much the latent weights differ between the two halves, and $\Delta_{AB}^{\mathrm{sd}}$ measures how much the standard deviations over time of the two half-averages differ. The error term $\varepsilon_T^{\mathrm{max}}$ behaves as for the ATV estimator: it becomes $\varepsilon_T$ when the argument of the square root is bounded away from zero uniformly in $T$, but for units with weak signal it is of order $T^{-1/4}$ at small $T$.

The SHSB and ATV estimators are closely related. When $\mathcal{S}$ contains all possible splits, the average of $\Delta_{AB}^{w}$ over splits is exactly $\frac{1}{n-1}\sum_k\widehat{\mathrm{Var}}_\ell\left[w_{ik}\right]$ (Lemma~\ref{lem:split-average}). The averaged numerators therefore coincide with the numerator of \eqref{eq:ATV main text}, and both estimators share the same $\mathcal{O}\left(1/n\right)$ underestimation due to the heterogeneity of the signal across trials. They differ only through $\Delta_{AB}^{\mathrm{sd}}$ in the denominators.

\subsection{Noise Ceiling for Single-Trial Responses}\label{sec:results_th_single}
The true noise ceiling $NC_{i,\ell}^{\mathrm{single}}$, derived in Appendix~\ref{sec:true_noise_ceiling_single}, is given by
\begingroup\makeatletter\def\f@size{9}\check@mathfonts
\begin{equation}
    NC^{\mathrm{single}}_{i,\ell}=\sqrt{\frac{\sum_k\left(w_{ik}^{(\ell)}\right)^2}{\sum_k\left(w_{ik}^{(\ell)}\right)^2+\sigma_\ell^2}}+\varepsilon_T.
\end{equation}
\endgroup
Unlike the trial-averaged case, it does not depend on the number of trials $n$, it is determined solely by the signal-to-noise ratio of the individual recording unit.\par
\paragraph{LOOC and TTMC estimators} The expressions for the LOOC and TTMC estimators, detailed in Appendices~\ref{sec:LOOC} and \ref{sec:TTMC}, are
\begingroup\makeatletter\def\f@size{9}\check@mathfonts
\begin{align}
    \widehat{NC}^{\mathrm{LOOC}}_{i,\ell}&=NC^{\mathrm{single}}_{i,\ell}\cdot\frac{\sum_kw_{ik}^{(\ell)}\bar{w}_{ik}^{(-\ell)}}{\sqrt{\left(\sum_k\left(w_{ik}^{(\ell)}\right)^2\right)\left(\sum_k\left(\bar{w}_{ik}^{(-\ell)}\right)^2+\frac{\bar{\sigma}^2_{(-\ell)}}{n-1}\right)}}+\varepsilon_T\nonumber,\\
    \widehat{NC}^{\mathrm{TTMC}}_{i,\ell}&=NC^{\mathrm{single}}_{i,\ell}\cdot\frac{\sum_kw_{ik}^{(\ell)}\bar{w}_{ik}+\frac{\sigma_\ell^2}{n}}{\sqrt{\left(\sum_k\left(w_{ik}^{(\ell)}\right)^2\right)\left(\sum_k\bar{w}_{ik}^2+\frac{\bar{\sigma}^2}{n}\right)}}+\varepsilon_T.
\end{align}
\endgroup
As $n$ becomes large, both expressions approach $NC_{i,\ell}^{\mathrm{single}}\cdot \mathrm{cos}_k\left(w^{(\ell)}_i,\overline{w}_i\right)+\varepsilon_T$, with $\mathrm{cos}_k(\cdot,\cdot)$ the cosine similarity across $k$.
However, in the heterogeneous setting, this limit is strictly lower than $NC_i^{\mathrm{single}}$. 
Notably, this bias arises because encoding models produce a separate prediction for each trial.
Other approaches that make a single prediction shared across all trials, such as Representational Similarity Analysis (RSA)~\citep{kriegeskorteRepresentationalSimilarityAnalysis2008}, are not affected by this bias.
In the homogeneous regime, both estimators converge to the true noise ceiling $NC_{i,\ell}^{\mathrm{single}}$ as $n\rightarrow\infty$, up to $\varepsilon_T$. The LOOC estimator always approaches it from below. The TTMC estimator approaches it from above as long as the noise variance of trial $\ell$ is at least half the average noise variance across trials, $\sigma_\ell^2\geq\bar{\sigma}^2/2$.
\paragraph{LOOP estimator} LOOP does not admit a closed form in terms of the weights, but its behavior can be described geometrically (Appendix~\ref{sec:LOOP}). When $T$ is large compared to the number $p=m(n-1)$ of predictors, $\widehat{NC}^{\mathrm{LOOP}}_{i,\ell}=NC^{\mathrm{single}}_{i,\ell}\cos\alpha^{(\ell)}$, where $\alpha^{(\ell)}$ is the angle between $\mathbf{v}^{*(\ell)}_i$ and the space spanned by the predictors. LOOP is therefore a lower bound on the single-trial noise ceiling, at least as large as LOOC in the large $T/p$ limit. As $n$ grows while keeping $T/p$ large, $\alpha^{(\ell)}$ decreases and LOOP eventually reaches the true noise ceiling, even in the heterogeneous regime. In practice, however, the number of time points $T$ is limited, and the generalization error of the fitted model grows with $p$, so that LOOP can in practice fall below LOOC. Its dependence on $n$ therefore combines an increasing and a decreasing term, and has no simple universal form. In particular, the exponential form $1-e^{-n/\tau_0}$ assumes a fast saturation and ignores the generalization error, both of which lead to underestimating the noise ceiling, which may explain why encoding model scores are frequently reported to exceed noise ceilings estimated with this method~\citep{awInstructiontuningAlignsLLMs2024}.

\subsection{Experimental Results} \label{sec:experimental_results}
In this section, we evaluate the noise ceiling estimators both on synthetic brain recordings and real fMRI and EEG data. The synthetic dataset, built as a specific setting of our general mathematical framework, provides a quantitative comparison between the true noise ceiling, the closed-form expressions derived in Sec.~\ref{sec:results_th_mean} and \ref{sec:results_th_single}, and the empirical estimates. The real datasets provide a qualitative validation of the theoretical and synthetic results, across different modalities and experimental settings. All results except LOOP were computed on a single CPU in less than a day in total. LOOP was computed on a single NVIDIA A100 GPU (80 GB) in approximately one day. The code used to generate all figures is available at \url{https://github.com/AudeMaier/noise-ceiling}.

\begin{figure}
     \centering
     \includegraphics[width=0.8\linewidth]{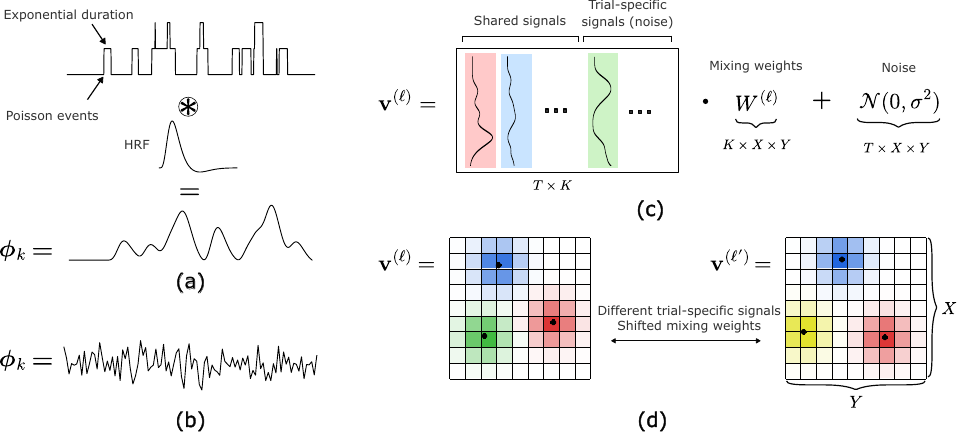}
     \caption{\textbf{Signal and trial generation.} (a) To mimic the fMRI BOLD response to a continuous stimulus, signals are generated by convolving an event time series with the hemodynamic response function (HRF). (b) To mimic the response to a set of randomly ordered discrete stimuli, signals are generated as i.i.d. Gaussian processes. (c) A trial is generated by multiplying shared and trial-specific signals with trial-specific mixing weights, and adding Gaussian noise. (d) Two generated trials differ by their trial-specific signals and their (potentially) misaligned mixing weights. More details about the generative process are provided in Sec. \ref{sec:experimental_results}.}
     \label{fig:signal_trial_generation}
\end{figure}

\vspace{-7pt}

\paragraph{fMRI and EEG datasets} We evaluate the noise ceiling estimators on three datasets. In each case, we describe the data in terms of $n$ trials, $m$ recording units, and a stimulus sequence of length $T$.
\begin{itemize}
\vspace{-6pt}
\item \textbf{THINGS fMRI1:} an fMRI dataset \citep{hebartTHINGSdataMultimodalCollection2023} in which 3 participants viewed sequences of images from the THINGS database \citep{hebartTHINGSDatabase18542019} while being scanned on a 3T MRI scanner. Each participant underwent 12 sessions, over which 100 images were repeated. We use the response estimates ($\beta$ values) obtained from a general linear model (GLM) for voxels in V1 of the first participant. The resulting dataset contains $n=12$ trials from a single participant, $m=1049$ units (voxels in V1), and $T=100$ images.

\item \textbf{THINGS EEG2:} an EEG dataset \citep{giffordLargeRichEEG2022} in which 10 participants viewed sequences of images from the THINGS database \citep{hebartTHINGSDatabase18542019}. The preprocessed data contain signals from 17 channels, sampled at 100 Hz. Each subject underwent 4 sessions, each containing 20 repetitions of the same 200 images.
We only use data from the first participant, and treat every combination of channel and time point as a separate unit. The resulting dataset contains $n=80$ trials from a single participant, $m=1700$ units (channels $\times$ time points), and $T=200$ images.

\item \textbf{Sherlock dataset:} an fMRI dataset  \citep{chenSharedMemoriesReveal2017} in which 16 participants watched the first episode of the BBC television series "Sherlock" (48 minutes), while being scanned on a 3T MRI scanner. We preprocessed the data using fMRIPrep \citep{estebanFMRIPrepRobustPreprocessing2019} and mapped them into MNI space. We isolated V1 and IT brain regions using the Harvard-Oxford atlas.
In summary, the dataset contains $n=16$ inter-subject trials, with $m=623$ units (voxels, 296 in V1 and 327 in IT), and a sequence length of $T=1976$ time points.

\end{itemize}
\vspace{-6pt}
The Sherlock dataset is clearly in the heterogeneous regime as its trials are different subjects, whose brains differ anatomically and functionally. The THINGS fMRI1 and EEG2 datasets are more ambiguous. Their trials are repetitions in the same participant, however they are spread across several sessions, which can introduce some heterogeneity (e.g. different electrode positions or internal state from one session to the next). We nevertheless assume homogeneity when generating the synthetic data that mimic these two datasets, and discuss the consequences of a possible heterogeneity in Sec.~\ref{sec:discussion}.

\paragraph{Synthetic brain recordings} We construct synthetic datasets consisting of $n$ trials of $m = 10 \times 10$ voxels arranged on a 2D grid, generated from the signal model of \eqref{eq: v}. The trial generation process is illustrated in Fig.~\ref{fig:signal_trial_generation}. The shared signals $\boldsymbol{\phi}_k$ can be constructed in two different ways. To mimic fMRI BOLD signals corresponding to one long continuous stimulus (Sherlock dataset), the signals $\boldsymbol\phi_k$ are obtained by simulating neural events according to a Poisson process ($\lambda=0.1$) with exponentially distributed durations ($\mu=5$), producing a box-shaped time series. This time series is then convolved with a double-gamma hemodynamic response function (HRF) to produce BOLD-like signals with time resolution $1.5\mathrm{s}$. For experimental setups where subjects are shown a set of discrete stimuli reshuffled in each trial (THINGS fMRI1 and THINGS EEG2), the signals $\boldsymbol{\phi_k}$ are generated as i.i.d. Gaussian processes. The noise term $\boldsymbol{\eta}_i^{(\ell)}$ consists of two components. The first follows the same structure as the signal term: it is a linear combination of trial-specific signals $\tilde{\boldsymbol{\phi}}_{k'}^{(\ell)}$, generated independently for each trial using the same procedure as the shared signals, with trial-specific mixing weights $\tilde{w}_{ik'}^{(\ell)}$. The fraction of shared signals among all (shared and trial-specific) signals is controlled by the parameter $\rho_{\mathrm{shared}}=K/(K+K')$. The second noise term is white Gaussian noise whose standard deviation varies across voxels as $\sigma_i = \sigma_0\, e^{sZ_i}$ with $Z_i \sim \mathcal{N}(0,1)$, so that $\sigma_0$ is the median noise level and $s$ controls its spread across voxels. The spatial structure of the mixing weights is inspired by \cite{erhardtSimTBSimulationToolbox2012}: for each signal $\boldsymbol\phi_k$, a center coordinate $(x_k, y_k)$ is drawn uniformly on the grid, and the weight $w_{ik}$ of voxel $i$ is set to a Gaussian kernel of standard deviation $\sigma_w$ evaluated at the distance between the voxel center and $(x_k, y_k)$. In the homogeneous regime, the coordinates are fixed across trials, yielding constant weights $w_{ik}$. In the heterogeneous regime, the center coordinates are resampled for each trial as $(x_k^{(\ell)}, y_k^{(\ell)}) = (x_k, y_k) + (\delta_x, \delta_y)$, where $(\delta_x, \delta_y) \sim \mathcal{U}([-\Delta, \Delta]^2)$ is drawn independently for each trial, producing trial-specific weights $w_{ik}^{(\ell)}$ that vary around a common mean. The parameters $\rho_\mathrm{shared}$, $\sigma_0$, $s$, $\sigma_w$, and $\Delta$ can be varied to model different experimental settings and brain areas.

\begin{figure}[h]
    \centering
    \includegraphics[width=0.79\linewidth]{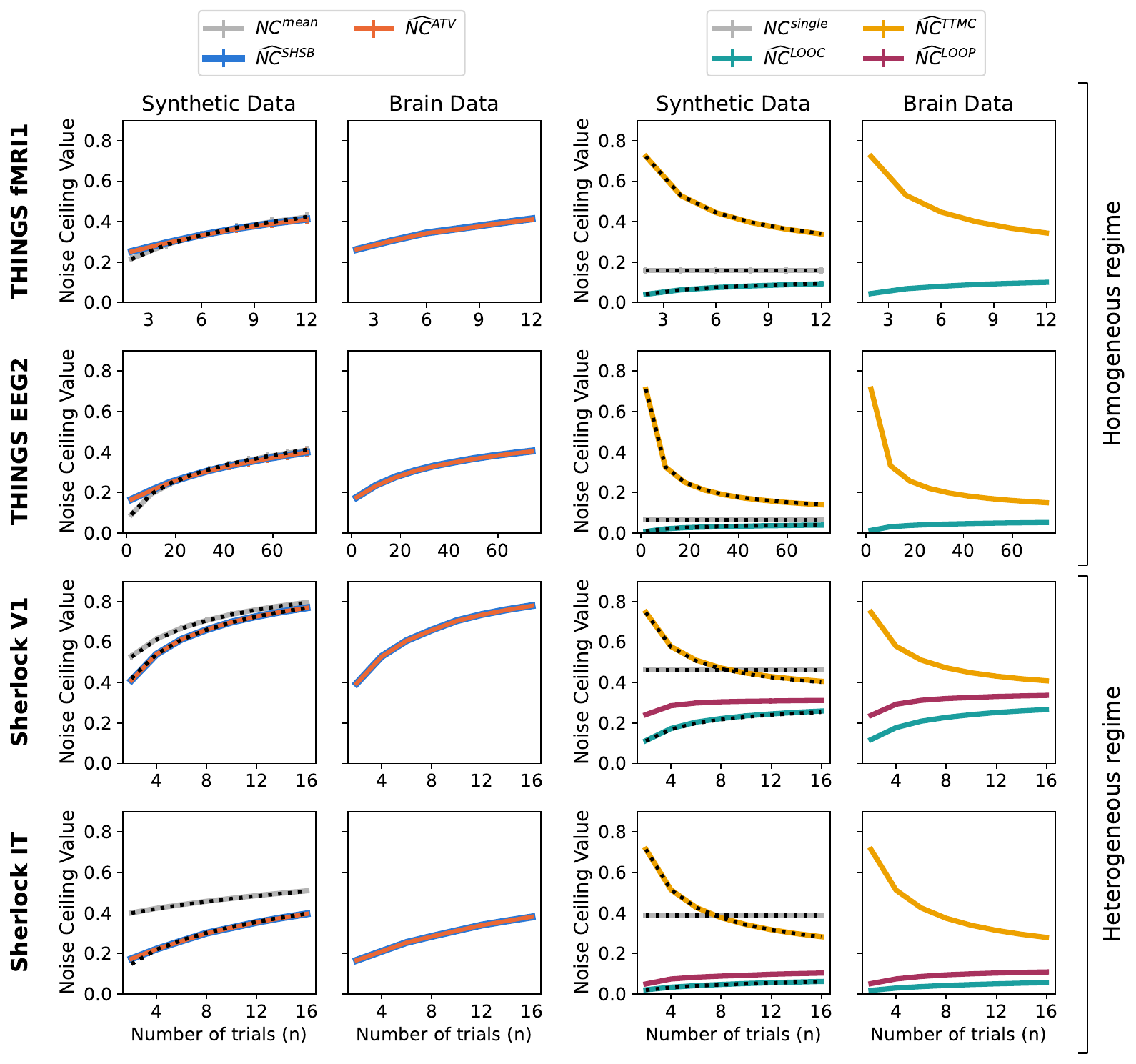}
    \caption{
    \textbf{Noise ceiling estimators on synthetic and brain data.} Each row corresponds to one dataset from Sec.~\ref{sec:experimental_results}. The left column shows the trial-averaged noise ceiling $NC^{\mathrm{mean}}$ and its estimators, and the right column shows the single-trial noise ceiling $NC^{\mathrm{single}}$ and its estimators, as a function of the number of trials $n$. Within each column, synthetic data and brain data are shown side by side, all values are averaged across units $i=1,\dots,m$. The true noise ceilings (gray) are known only for the synthetic data. For the synthetic data, solid lines show the mean over 10 simulated datasets, with error bars giving asymptotic 95\% confidence intervals (often smaller than the line width), and dotted black lines show the leading term of the corresponding closed form from Sec.~\ref{sec:results_th_mean},~\ref{sec:results_th_single}. The synthetic data are generated in the homogeneous regime ($\Delta=0$) in the top two rows and in the heterogeneous regime ($\Delta>0$) in the bottom two rows, with parameters chosen to mimic the brain data of each row (all parameters are given in Table~\ref{tab:sim_params}, Appendix~\ref{app:synthetic_param}). For each $n$, estimates are averaged over 50 random subsets of $n$ trials. LOOP, which has no closed form, is shown only for Sherlock and its synthetic counterpart. THINGS fMRI1 and THINGS EEG2 have too few stimuli (100 and 200) to fit the ridge regression meaningfully, and in practice LOOP is used for across-subject repetitions.
}
    \label{fig:results}
    \vspace{-6pt}
\end{figure}

\paragraph{Finite-$T$ bias} Fig.~\ref{fig:results} shows the estimators as a function of the number of trials $n$ for the three datasets and for synthetic datasets designed to mimic them. For the synthetic data, the empirical values can be compared with the leading term of their closed-form expressions; the gap between a solid line and the corresponding dotted black line estimates the bias $\mathbb{E}[\varepsilon_T]$ of that estimator. The two lines match closely, except for ATV and SHSB at small $n$, which exhibit a positive bias when the noise ceiling is low. This is the transient regime described in Sec.~\ref{sec:results_th_mean}, which arises for voxels with low true noise ceiling. As the true noise ceiling increases with $n$, the bias disappears for larger $n$. This phenomenon can be studied further by plotting the estimators as a function of $T$ at fixed $n$, as done on the Sherlock dataset in Appendix~\ref{app:finite-T-sherlock}.
\paragraph{Homogeneous regime}
Looking at the top two rows of Fig.~\ref{fig:results}, the leading terms of ATV and SHSB both overlap with $NC^{\mathrm{mean}}$, up to their finite-$T$ biases. Here $\Delta^{\mathrm{sd}}_{AB}=0$ because all trials have the same noise variance in the synthetic data, making ATV and SHSB equal. In the brain data, ATV and SHSB also nearly coincide, showing that $\Delta^{\mathrm{sd}}_{AB}$ is small in practice. For the trial-specific noise ceiling, LOOC and TTMC correctly approach $NC^{\mathrm{single}}$ as $n$ grows, from below and from above, respectively.
\paragraph{Heterogeneous regime}
In the bottom two rows of Fig.~\ref{fig:results}, ATV and SHSB underestimate $NC^{\mathrm{mean}}$ by a term of order $1/n$, consistently with their closed forms in the heterogeneous regime. This term grows with the variance of the weights across trials, which the synthetic data control through $\Delta$ and $\sigma_w$. The trial-specific estimators, LOOC and TTMC, converge to a common limit below $NC^{\mathrm{single}}$. TTMC is therefore no longer an upper bound as it falls below $NC^{\mathrm{single}}$ from $n\approx 8$ in both synthetic settings. LOOP lies above LOOC in both brain regions. In the synthetic data, however, it saturates far below $NC^{\mathrm{single}}$, so extrapolating it to large $n$ with an exponential fit would still underestimate the noise ceiling. The true noise ceilings of the synthetic data should not be read as estimates of true ceilings of the brain data. Many parameter combinations reproduce the same estimator curves with different true noise ceilings. $NC^{\mathrm{mean}}$ can in principle lie anywhere between ATV and 1, depending on the level of heterogeneity. The true noise ceilings of the synthetic data are only one plausible example of where those of the brain data could lie. Similarly, LOOP lying above LOOC is not evidence of heterogeneity. In the homogeneous regime, when the signal of neighboring units is correlated, LOOP can combine many units to average out more of their noise than LOOC.
\vspace{-2pt}
\section{Discussion}\label{sec:discussion}
\vspace{-2pt}
In this work, we introduced a unified mathematical framework for brain recordings across modalities and used it to analyze the statistical properties of noise ceiling estimators across experimental settings. We identified two central regimes: the homogeneous regime, in which all trials express the latent signals with the same mixing weights, and the heterogeneous regime, in which these weights vary across trials. In the homogeneous regime, ATV recovers $NC^{\mathrm{mean}}$ up to finite-$T$ error terms. SHSB does so for even $n$ when all trials have the same noise level. LOOC and TTMC converge to $NC^{\mathrm{single}}$ as $n$ grows. In the heterogeneous regime, ATV and SHSB underestimate $NC^{\mathrm{mean}}$ by a term of order $1/n$, and LOOC and TTMC converge to a limit below $NC^{\mathrm{single}}$. LOOP is the only estimator that can converge to $NC^{\mathrm{single}}$ in this regime, but for realistic values of $T$ its generalization error grows with the number of predictors, and in our synthetic data it saturates below $NC^{\mathrm{single}}$.
Heterogeneity is typically important when trials correspond to different subjects, but it can also arise between sessions of the same subject, and even within a session: the response to a stimulus seen for the first time can differ from the response to its later presentations, once it has become familiar. How much heterogeneity these factors introduce is an important question for future work. If it is non-negligible, the estimators would underestimate the true noise ceiling also for datasets with repeated presentations in the same participant, such as THINGS-fMRI1 and THINGS-EEG2, although for ATV and SHSB this underestimation decreases with the number of repetitions.\par
The analytical results rely on four assumptions (Assumptions~\ref{ass:independent-trials}--\ref{ass:summable-fourth-cumulants}). Noise that is correlated between trials violates the first one and inflates all estimators, which then treat the shared noise as signal. Stimulus-locked non-neuronal responses, such as changes in heart rate or gaze at specific events, do not violate it. They are reproducible across trials and therefore part of the signal, which a model of the stimulus could in principle predict. For trials recorded in the same run, shared noise can come from slow drifts, which are typically removed by high-pass filtering. The second assumption, boundedness of the signals, is trivial for any physical recording. The last two require that the noise and its squared amplitude have short memory, so that the error terms vanish as $T$ grows. Slow drifts and persistent oscillations violate them and can lead to errors that decay slowly or not at all. They can be largely removed by high-pass filtering and nuisance regression, and the remaining memory can be evaluated empirically by observing the decay of the autocorrelation of the signals and of their squares.\par
Table~\ref{tab:recommendations} summarizes which estimator to use depending on the regime and on whether the model is evaluated on trial-averaged or single-trial responses. In the homogeneous regime, the trial-averaged noise ceiling can be estimated without systematic error, whereas in the heterogeneous regime only lower bounds are available. In particular, TTMC should not be used as an upper bound in the heterogeneous regime. For single trials, variants of LOOP that reduce the generalization error at the cost of a larger angle $\alpha^{(\ell)}$ can also be used, for instance predicting each unit only from the same unit in the other trials, or from the average of the other trials. Taking, for each unit, the highest of LOOC, LOOP and its variants provides a tighter lower bound on $NC^{\mathrm{single}}$, provided that the estimator is selected on a validation set, to avoid biasing the estimate upwards. When only a lower bound on the noise ceiling is available, we recommend reporting brain scores as an interval: its lower end is the unnormalized brain score and its upper end is the brain score normalized by the lower bound, clipped at $1$. A checklist of common pitfalls when computing noise ceilings is provided in Appendix~\ref{sec:checklist}.
\begin{table}[t]
\centering\small
\caption{Recommended noise ceiling estimators.}
\label{tab:recommendations}
\begin{tabular}{@{}p{0.15\linewidth}p{0.36\linewidth}p{0.41\linewidth}@{}}
\toprule
& Trial-averaged ($NC^{\mathrm{mean}}$) & Single-trial ($NC^{\mathrm{single}}$) \\
\midrule
Homogeneous & ATV: unbiased up to finite-$T$ effects & LOOC: lower bound,
TTMC: upper bound$^\dagger$\vspace{3pt} \\
Heterogeneous & ATV: lower bound & Highest of LOOC, LOOP and LOOP variants (selected on a validation set): lower bound\\
\bottomrule
\end{tabular}
\vspace{2pt}
{\footnotesize $^\dagger$Requires the noise variance of
the trial to be at least half the average noise variance, $\sigma_\ell^2 \ge
\bar\sigma^2/2$.}
\vspace{-22pt}
\end{table}
\vspace{-3pt}
\paragraph{Limitations}\label{par:limitations}
While our framework provides a unified theoretical grounding for noise ceiling estimation, several theoretical and empirical limitations remain. First, the reproducible response is formally defined to include everything that is predictable given the attributes that are fixed within a trial. This choice matches the standard use of encoding models as predictors of the recorded responses. If the goal is instead to assess whether a model is brain-like, i.e. whether it captures what is common to brains, only the latent signals shared by most subjects should be considered, which would lower the true noise ceiling. The main conclusions of the paper would, however, still hold as differences in how the subjects mix these shared signals would still exist, resulting in downward biases of the estimators. Additionally, although we provide initial theoretical analysis for Leave-One-Out Prediction (LOOP), a complete theoretical treatment remains open. Empirically, our evaluation is restricted to fMRI and EEG datasets using visual (THINGS fMRI and THINGS EEG) or audiovisual (Sherlock) stimuli, leaving other modalities (invasive electrophysiology, magnetoencephalography) and other cognitive domains (language processing, motor control) unexamined. Finally, our empirical validation relies on consistency checks of the closed-form expressions using synthetic data generated under a particular case of our theoretical framework, and qualitative comparisons across three datasets, rather than an exhaustive quantitative benchmark across diverse experimental and synthetic setups.

\section*{Acknowledgment}
This work was supported by the Swiss National Science Foundation grant 200021-236436.
\bibliography{references_aude}

\newpage
\appendix
\section{Noise ceiling methods used in the literature}
{\footnotesize
\renewcommand{\arraystretch}{1.35}

\begin{longtable}{|p{1.7cm}|p{3cm}|p{1.5cm}|p{1.5cm}|p{1.5cm}|p{1.5cm}|}

\caption{Summary of noise ceiling methods used in the neuroimaging literature, grouped by method type. For each paper, we report the stimulus modality, whether a General Linear Model (GLM) was used to extract per-voxel beta values, whether the dataset included multiple repetitions of the same stimulus within participants, and whether the model was fit to predict a participant average (as opposed to individual participants separately).} \label{table} \\

\hline
\textbf{Noise ceiling method} & \textbf{Papers} & \textbf{Modality} & \textbf{GLM} & \textbf{Repetitions} & \textbf{Predict average} \\ \hline

% --- SB SECTION ---
\multirow{3}{*}{\rotatebox[origin=c]{0}{SHSB}} 
 & \citet{tangDiversePerceptualRepresentations2025} & Video & No & Yes and No & Yes \\ \cline{2-6}
 & \citet{garciaModelingDynamicSocial2025} & Video & Yes & Yes & No \\ \cline{2-6}
 & \citet{tuckuteManyNotAll2023} & Audio & Yes and No & Yes & No \\ \cline{2-6}
 & \citet{lebelNaturalLanguageFMRI2023} & Language & Yes & Yes & No \\ \cline{2-6}
 & \citet{moussaImprovingSemanticUnderstanding2025} & Language & No & Yes & No \\ \hline

% --- ATV SECTION ---
\multirow{7}{*}{\rotatebox[origin=c]{0}{ATV}} 
 & \citet{allenMassive7TFMRI2022} (NSD dataset) & Image & Yes & Yes & No \\ \cline{2-6}
 & \citet{khoslaHigherVisualAreas2026} (NSD dataset) & Image & Yes & Yes & No \\ \cline{2-6}
 & \citet{hebartTHINGSdataMultimodalCollection2023} & Image & Yes & Yes & No \\ \cline{2-6}
 & \citet{tuckuteDrivingSuppressingHuman2024} & Language & Yes & No & Yes \\ \cline{2-6}
 & \citet{antonelloScalingLawsLanguage2023} & Language & No & Yes & No \\ \hline

 % --- LOOC SECTION ---
\multirow{1}{*}{\rotatebox[origin=c]{0}{LOOC}} 
 & \citet{honarmandInducingDyslexiaVision2026} & Language & No & No & No \\ \hline
 
% --- TTM CORRELATION SECTION ---
\multirow{2}{*}{\rotatebox[origin=c]{0}{\small TTMC}} 
 & \citet{rahimiExplanationsLargeLanguage2025} & Language & No & No & No \\ \cline{2-6}
 & \citet{kumarSharedFunctionalSpecialization2024} & Audio & No & No & No \\ \hline

% --- LOO PREDICTION SECTION ---
\multirow{10}{*}{\rotatebox[origin=c]{0}{LOOP}} 
 & \citet{schrimpfNeuralArchitectureLanguage2021} & Language & No & Yes & No \\ \cline{2-6}
 & \citet{merlinLanguageModelsBrains2024} & Language & No & Yes & No \\ \cline{2-6}
 & \citet{kaufLexicalSemanticContentNot2024} & Language & Yes & Yes & No \\ \cline{2-6}
 & \citet{hosseiniArtificialNeuralNetwork2024a} & Language & No & Yes & No \\ \cline{2-6}
 & \citet{linVectorsLargerLanguage2025} & Language & No & Yes & No \\ \cline{2-6}
 & \citet{awTrainingLanguageModels2022} & Language & No & No & No \\ \cline{2-6}
 & \citet{awInstructiontuningAlignsLLMs2024} & Language & No & Yes and No & No \\ \cline{2-6}
 & \citet{ootaSpeechLanguageModels2024} & Language & No & No & No \\ \cline{2-6}
 & \citet{alkhamissiBrainLikeLanguageProcessing2024} & Language & No & Yes and No & Yes and No \\ \cline{2-6}
 & \citet{gaoIncreasingAlignmentLarge2025} & Language & No & No & No \\ \hline

\end{longtable}
}

\section{Mathematical derivations}\label{app:mathematical_derivations}
\subsection{Statement on AI use}\label{app:ai_statement}
The proofs of Lemma~\ref{lem:delta-method}, Theorem~\ref{thm:general-limit-theorem}, Corollary~\ref{cor:correlation-of-trial-averages}, Lemma~\ref{lem:thresholded-functions}, Proposition~\ref{prop:ATV}, and Proposition~\ref{prop:shsb} were developed with AI assistance. An LLM (Claude Opus 5 and 5.5) was used to brainstorm proof skeletons and draft the proofs in an iterative process with the authors. The authors verified every step of these proofs, checking each argument and redoing all computations by hand. Additionally, the authors detailed, clarified or simplified the arguments where needed. All other results were derived independently by the authors. For these, the LLM was used for writing assistance and to check the derivations. The authors take full responsibility for the correctness of all statements and proofs.
\subsection{Large $T$ Convergence Rate}\label{sec:large_T_convergence}
In this section, we fix one recording unit and omit the index $i$ for more readability. For each trial $\ell$, we define the corresponding brain recording as a stochastic process $\{v_{t}^{(\ell)}\}_{t>0}$. Each trial carries attributes $\theta^{(\ell)}$ (e.g. subject identity, experimental conditions). All results are stated conditionally on the attributes $\theta=\left(\theta^{(1)},...,\theta^{(n)}\right)$, i.e. all expectations $\mathbb{E}\left[\cdot\right]$, variances $\mathrm{Var}\left(\cdot\right)$ and covariances $\mathrm{Cov}\left(\cdot\right)$ are implicitly conditional on $\theta$.
\begin{definition}[signal and noise]\label{def:signal-and-noise} The noise-free signal and the noise of trial $\ell$ are defined as
\begin{equation}
    v_t^{*\left(\ell\right)}:=\mathbb{E}\left[v_t^{(\ell)}\right],\qquad \eta_t^{(\ell)}:=v_t^{\left(\ell\right)} - v_t^{*\left(\ell\right)},
\end{equation}
so that $\mathbb{E}[\eta^{(\ell)}_t]=0$ by construction. The signal $v^{*(\ell)}_t$ is deterministic (given $\theta$) and all randomness sits in the noise. We write
\begin{equation}
  \gamma_\ell(t,s) := \mathrm{Cov}\big(\eta^{(\ell)}_t,\eta^{(\ell)}_s\big)
\end{equation}
for the autocovariance of the noise of trial $\ell$, and
\begin{equation}\label{eq:sigma}
  \sigma_\ell^2 := \frac1T\sum_{t=1}^T \mathrm{Var}\big(\eta^{(\ell)}_t\big)=\frac1T\sum_{t=1}^T\gamma_\ell(t,t)
\end{equation}
for the noise variance of trial $\ell$, averaged over time.
\end{definition}
\begin{definition}[time averages]\label{def:time-averages} For finite sequences $\mathbf{x}=\{x_t\}_{t=1,...,T}$, $\mathbf{y}=\{y_t\}_{t=1,...,T}$ we write
\begin{align}
    \langle\mathbf{x}\rangle:=\frac{1}{T}\sum_{t=1}^Tx_t,&\qquad \widehat{\mathrm{Cov}}_t\left(\mathbf{x},\mathbf{y}\right):=\langle\mathbf{xy}\rangle - \langle\mathbf{x}\rangle \langle\mathbf{y}\rangle,\\
    \widehat{\mathrm{Var}}_t\left(\mathbf{x}\right):=\widehat{\mathrm{Cov}}_t\left(\mathbf{x},\mathbf{x}\right),&\qquad \widehat{\mathrm{Corr}}_t\left(\mathbf{x},\mathbf{y}\right):=\frac{\widehat{\mathrm{Cov}}_t\left(\mathbf{x},\mathbf{y}\right)}{\sqrt{\widehat{\mathrm{Var}}_t\left(\mathbf{x}\right)\widehat{\mathrm{Var}}_t\left(\mathbf{y}\right)}},
\end{align}
where $\mathbf{xy}$ is the elementwise product.
\end{definition}
\begin{definition}[averages over subsets of trials]\label{def:averages-over-subsets-of} For a non-empty subset of trials $A\subseteq[n]$ we write
\begin{equation}
    \mathbf{v}^A:=\frac{1}{|A|}\sum_{\ell\in A}\mathbf{v}^{(\ell)},\qquad \mathbf{v}^{*A}:=\frac{1}{|A|}\sum_{\ell\in A}\mathbf{v}^{*(\ell)},\qquad \boldsymbol{\eta}^A:=\frac{1}{|A|}\sum_{\ell\in A}\boldsymbol{\eta}^{(\ell)}
\end{equation}
so that $\mathbf{v}^A=\mathbf{v}^{*A}+\boldsymbol{\eta}^{A}$. Additionally, for two subsets $A,B\subseteq[n]$ we define their \emph{shared noise} as
\begin{equation}
    s\left(A,B\right):=\frac{1}{|A||B|}\sum_{\ell\in A\cap B} \sigma_\ell^2 .
\end{equation}
\end{definition}

We now list all assumptions needed for the derivations of our results below.
 
\begin{assumption}[independent trials]\label{ass:independent-trials} The noise processes $\boldsymbol{\eta}^{\left(1\right)},...,\boldsymbol{\eta}^{\left(n\right)}$ are mutually independent (conditionally on $\theta$).
\end{assumption}
\begin{assumption}[bounded signals]\label{ass:bounded-signals} There exists a constant $M<\infty$ such that $\mathrm{sup}_{\ell\in[n]}\mathrm{sup}_{t>0}\left|v^{*(\ell)}_t\right|<M$.
\end{assumption}
\begin{assumption}[short-range temporal correlations]\label{ass:short-range-temporal-correlations} There exists a constant $C<\infty$ such that $\mathrm{sup}_{\ell\in[n]}\mathrm{sup}_{t>0}\sum_{s=1}^{\infty}\left|\gamma_\ell\left(t,s\right)\right|<C$.
\end{assumption}
\begin{assumption}[short-range correlation of the squared noise]\label{ass:summable-fourth-cumulants} For every trial $\ell$, $\mathbb{E}\left[\left(\eta^{(\ell)}_t\right)^4\right]<\infty$ for all $t$, and there exists a constant $K_4<\infty$ such that $\mathrm{sup}_{\ell\in[n]}\mathrm{sup}_{t>0}\sum_{s=1}^{\infty}\left|\mathrm{Cov}\left(\left(\eta^{(\ell)}_t\right)^2,\left(\eta^{(\ell)}_s\right)^2\right)\right|<K_4$.
\end{assumption}

 \paragraph{Error terms.} Throughout, $\varepsilon_T$ denotes a random error term whose bias and mean squared error satisfy $\left|\mathbb{E}\left[\varepsilon_T\right]\right|\leq c/T$ and $\mathbb{E}\left[\varepsilon_T^2\right]\leq c/T$, and $\varepsilon_T^{\mathrm{max}}$ a random error term satisfying $\left|\mathbb{E}\left[\varepsilon_T^{\mathrm{max}}\right]\right|\leq(c/T)^{1/4}$ and $\mathbb{E}\left[\left(\varepsilon_T^{\mathrm{max}}\right)^2\right]\leq(c/T)^{1/2}$, for a constant $c$ that does not depend on $T$, nor $n$. Each result specifies the quantities on which $c$ may depend.
\begin{lemma}[noise statistics]\label{lem:noise-statistics} Let $A,B\subseteq[n]$ be non-empty. Under Assumptions~\ref{ass:independent-trials}--\ref{ass:summable-fourth-cumulants},
\begin{enumerate}[(a)]
    \item \textbf{(time average of the noise)} $\mathbb{E}\left[\langle\boldsymbol{\eta}^A\rangle\right]=0$ and ${\mathrm{Var}}\left(\langle\boldsymbol{\eta}^A\rangle\right)\leq\frac{C}{|A|T}$
    \item \textbf{(signal-noise covariance)} $\mathbb{E}\left[\widehat{\mathrm{Cov}}_t\left(\mathbf{v}^{*A},\boldsymbol{\eta}^B\right)\right]=0$ and $\mathrm{Var}\left(\widehat{\mathrm{Cov}}_t\left(\mathbf{v}^{*A},\boldsymbol{\eta}^B\right)\right)\leq\frac{4M^2C}{|B|T}$
    \item \textbf{(noise-noise covariance)} Write $\widehat{\mathrm{Cov}}_t\left(\boldsymbol{\eta}^{A},\boldsymbol{\eta}^B\right)=s(A,B)+Y_{AB}+W_{AB}$ with
    \begin{equation}
        Y_{AB}:=\langle\boldsymbol{\eta}^A\boldsymbol{\eta}^B\rangle-s(A,B),\qquad W_{AB}:=-\langle\boldsymbol{\eta}^A\rangle\langle\boldsymbol{\eta}^B\rangle.
    \end{equation}
    Then $\mathbb{E}\left[Y_{AB}\right]=0$, $\mathbb{E}\left[Y_{AB}^2\right]\leq\frac{C^2+K_4}{T}$, and $\mathbb{E}\left[|W_{AB}|\right]\leq\frac{C}{T}$.
\end{enumerate}
\end{lemma}
\begin{proof}
We first compute two quantities that will be useful throughout the proof. By
Assumption~\ref{ass:independent-trials}, noise from different trials is uncorrelated, so for any $A$,
\begin{equation}\label{eq:avgcov}
  \mathrm{Cov}\big(\eta^A_t,\eta^A_s\big)=\frac1{|A|^2}\sum_{\ell,\ell'\in A}\mathrm{Cov}\big(\eta^{(\ell)}_t,\eta^{(\ell')}_s\big)
  =\frac1{|A|^2}\sum_{\ell\in A}\gamma_\ell(t,s),
\end{equation}
and therefore, by Assumption~\ref{ass:short-range-temporal-correlations}, for every $t$,
\begin{equation}\label{eq:avgsum}
  \sum_{s=1}^T\big|\mathrm{Cov}(\eta^A_t,\eta^A_s)\big|
  \le\frac1{|A|^2}\sum_{\ell\in A}\sum_{s=1}^\infty|\gamma_\ell(t,s)|\le\frac{C}{|A|}.
\end{equation}
\begin{enumerate}[(a)]
    \item Linearity of expectation implies $\mathbb{E}\left[\langle\boldsymbol{\eta}^A\rangle\right]=0$. Moreover,
    \begin{equation}
    \mathrm{Var}\big(\langle{\boldsymbol \eta^A}\rangle\big)=\frac1{T^2}\sum_{t=1}^T\sum_{s=1}^T\mathrm{Cov}(\eta^A_t,\eta^A_s)
  \le\frac1{T^2}\sum_{t=1}^T\frac{C}{|A|}=\frac{C}{|A|\,T}.
    \end{equation}
    \item Set $c_t:=v^{*A}_t-\langle{\mathbf{v}^{*A}}\rangle$. By linearity of the time average,
\begin{equation}
  \widehat{\mathrm{Cov}}_t(\mathbf{v}^{*A},\boldsymbol\eta^B)=\langle{\mathbf{v}^{*A}\boldsymbol{\eta}^B}\rangle-\langle{\mathbf{v}^{*A}}\rangle\langle{\boldsymbol\eta^B}\rangle
  =\frac1T\sum_{t=1}^T\big(v^{*A}_t-\langle{\mathbf{v}^{*A}}\rangle\big)\eta^B_t=\frac1T\sum_{t=1}^T c_t\,\eta^B_t ,
\end{equation}
which has mean zero because $c_t$ is deterministic. By Assumption~\ref{ass:bounded-signals},
$|v^{*A}_t|\le M$ and $|\langle{\mathbf{v}^{*A}}\rangle|\le M$, hence $|c_t|\le 2M$. Using~\eqref{eq:avgsum},
\begin{equation}
  \mathrm{Var}\Big(\frac1T\sum_{t}c_t\eta^B_t\Big)=\frac1{T^2}\sum_{t,s}c_tc_s\mathrm{Cov}(\eta^B_t,\eta^B_s)
  \le\frac{4M^2}{T^2}\sum_{t=1}^T\sum_{s=1}^T\big|\mathrm{Cov}(\eta^B_t,\eta^B_s)\big|
  \le\frac{4M^2C}{|B|\,T}.
\end{equation}
    \item We expand the product of averages into products of single trials,
\begin{equation}\label{eq:pairs}
  \langle{\boldsymbol\eta^A\boldsymbol\eta^B}\rangle=\frac1{|A|\,|B|}\sum_{\ell\in A}\sum_{\ell'\in B}\langle{\boldsymbol\eta^{(\ell)}\boldsymbol\eta^{(\ell')}}\rangle=\frac1{|A|\,|B|}\sum_{\ell\in A}\sum_{\ell'\in B}\left(\frac1T\sum_{t=1}^T\eta^{(\ell)}_t\eta^{(\ell')}_t\right).
\end{equation}
For $\ell\ne\ell'$, independence and zero mean give $\mathbb{E}\left[\langle{\boldsymbol\eta^{(\ell)}\boldsymbol\eta^{(\ell')}}\rangle\right]=0$. For
$\ell=\ell'$, $\mathbb{E}\left[ \langle{\boldsymbol\eta^{(\ell)}\boldsymbol\eta^{(\ell)}}\rangle\right]=\frac1T\sum_t\gamma_\ell(t,t)=\sigma_\ell^2$. Hence
$\mathbb{E}\left[ \langle{\boldsymbol\eta^{A}\boldsymbol\eta^{B}}\rangle\right]=\frac{1}{|A||B|}\sum_{\ell\in A\cap B}\sigma^2_\ell=s(A,B)$, which gives $\mathbb{E}[ Y_{AB}]=0$.\par
Still for $\ell\ne\ell'$, independence of the two trials gives
$\mathbb{E}[\eta^{(\ell)}_t\eta^{(\ell')}_t\eta^{(\ell)}_s\eta^{(\ell')}_s]
=\gamma_\ell(t,s)\gamma_{\ell'}(t,s)$. Note that $|\gamma_\ell(t,s)|\le \sqrt{\gamma_\ell(t,t)\gamma_\ell(s,s)}\le C$ by the Cauchy-Schwarz inequality, since $\gamma_\ell(t,t)\le\sum_{s}|\gamma_\ell(t,s)|\le C$. Therefore
\begin{equation}\label{eq:pair-offdiag}
  \mathrm{Var}\left(\langle{\boldsymbol\eta^{(\ell)}\boldsymbol\eta^{(\ell')}}\rangle\right)=\frac1{T^2}\sum_{t,s}\gamma_\ell(t,s)\gamma_{\ell'}(t,s)
  \le\frac1{T^2}\sum_{t=1}^T C\sum_{s=1}^\infty|\gamma_\ell(t,s)|\le\frac{C^2}{T}.
\end{equation}
For $\ell=\ell'$, Assumption~\ref{ass:summable-fourth-cumulants} gives
\begin{equation}\label{eq:pair-diag}
  \mathrm{Var}\left(\langle{\boldsymbol\eta^{(\ell)}\boldsymbol\eta^{(\ell)}}\rangle\right)=\frac1{T^2}\sum_{t,s}\mathrm{Cov}\left(\left(\eta^{(\ell)}_t\right)^2,\left(\eta^{(\ell)}_s\right)^2\right)\leq\frac{K_4}{T}.
\end{equation}

Using those results, and since $C^2,K_4\leq C^2+K_4$, we can compute
\begin{align}
    \mathrm{Var}\left(Y_{AB}\right)=\mathbb{E}\left[Y_{AB}^2\right]&=\mathbb{E}\left[\left(\frac{1}{|A||B|}\sum_{\ell\in A}\sum_{\ell'\in B}\left(\langle{\boldsymbol\eta^{(\ell)}\boldsymbol\eta^{(\ell')}}\rangle-\mathbb{E}\left[\langle{\boldsymbol\eta^{(\ell)}\boldsymbol\eta^{(\ell')}}\rangle\right]\right)\right)^2\right]\nonumber\\
    &\leq\left(\frac{1}{|A||B|}\sum_{\ell\in A}\sum_{\ell'\in B}\sqrt{\mathbb{E}\left[\left(\langle{\boldsymbol\eta^{(\ell)}\boldsymbol\eta^{(\ell')}}\rangle-\mathbb{E}\left[\langle{\boldsymbol\eta^{(\ell)}\boldsymbol\eta^{(\ell')}}\rangle\right]\right)^2\right]}\right)^2\nonumber\\
    &=\left(\frac{1}{|A||B|}\sum_{\ell\in A}\sum_{\ell'\in B}\sqrt{\mathrm{Var}\left(\langle{\boldsymbol\eta^{(\ell)}\boldsymbol\eta^{(\ell')}}\rangle\right)}\right)^2\leq\frac{C^2+K_4}{T},
\end{align}
where the first inequality used Minkowski's inequality.\par
Finally, using Cauchy-Schwarz, the fact that $\langle\boldsymbol\eta^{A}\rangle$ and $\langle\boldsymbol\eta^{B}\rangle$ have mean zero, and (a), we can write
\begin{equation}
    \mathbb{E}\left[|W_{AB}|\right]=\mathbb{E}\left[\left|\langle\boldsymbol\eta^{A}\rangle\langle\boldsymbol\eta^{B}\rangle\right|\right]\leq\sqrt{\mathbb{E}\left[\langle\boldsymbol\eta^{A}\rangle^2\right]\mathbb{E}\left[\langle\boldsymbol\eta^{B}\rangle^2\right]}=\sqrt{\mathrm{Var}\left(\langle\boldsymbol\eta^{A}\rangle\right)\mathrm{Var}\left(\langle\boldsymbol\eta^{B}\rangle\right)}\leq\frac{C}{T}.
\end{equation}
\end{enumerate}
 
\end{proof}
\begin{definition}[neighborhoods]\label{def:neighbourhoods} For $\mathbf{z}\in\mathbb{R}^d$ we write $|\mathbf{z}|$ for the Euclidean norm. For a set $\mathcal{K}\subset\mathbb{R}^d$ and $\delta>0$, we write $\mathcal{K}_\delta:=\left\{\mathbf{x}\in\mathbb{R}^d : |\mathbf{x}-\mathbf{z}|\leq\delta \text{ for some } \mathbf{z}\in\mathcal{K}\right\}$ for the set of points at distance at most $\delta$ from $\mathcal{K}$.
\end{definition}
 
\begin{lemma}[delta method]\label{lem:delta-method} Let $\{\hat{\mathbf{Z}}_T\}_{T>0}$ be random vectors and $\{\mathbf{z}_T\}_{T>0}$ deterministic vectors in $\mathbb{R}^d$ such that
\begin{equation}
    \hat{\mathbf{Z}}_T-\mathbf{z}_T=\mathbf{Y}_T+\mathbf{W}_T,\qquad \mathbb{E}\left[\mathbf{Y}_T\right]=0,\qquad \mathbb{E}\left[|\mathbf{Y}_T|^2\right]\leq\frac{c_1}{T},\qquad \mathbb{E}\left[|\mathbf{W}_T|\right]\leq\frac{c_2}{T},
\end{equation}
for some constants $c_1,c_2$. Let $\mathcal{K}$ be a compact set such that $\mathbf{z}_T\in\mathcal{K}$ for all $T$, let $\delta>0$, and let $F:\mathbb{R}^d\to\mathbb{R}$ be twice continuously differentiable on an open set containing $\mathcal{K}_\delta$. Since $\mathcal{K}_\delta$ is compact,
\begin{equation}
    L_1:=\mathrm{sup}_{\mathbf{x}\in\mathcal{K}_\delta}|\nabla F(\mathbf{x})|<\infty,\qquad L_2:=\mathrm{sup}_{\mathbf{x}\in\mathcal{K}_\delta}\left|\left|\mathbf{H}_{F}\left(\mathbf{x}\right)\right|\right|_{\mathrm{op}}<\infty,
\end{equation}
where $\mathbf{H}_F$ is the Hessian matrix of $F$.
Assume further that there exists a constant $U$ such that $|F(\hat{\mathbf{Z}}_T)|\leq U$ almost surely and $|F(\mathbf{z}_T)|\leq U$. Then
\begin{enumerate}[(a)]
    \item \textbf{(bias)} $\left|\mathbb{E}\left[F(\hat{\mathbf{Z}}_T)\right]-F(\mathbf{z}_T)\right|\leq\frac{c}{T}$,
    \item \textbf{(mean squared error)} $\mathbb{E}\left[\left(F(\hat{\mathbf{Z}}_T)-F(\mathbf{z}_T)\right)^2\right]\leq\frac{c}{T}$,
\end{enumerate}
where $c$ only depends on $c_1$, $c_2$, $\delta$, $U$, $L_1$, and $L_2$.
\end{lemma}
\begin{proof}
We drop the index $T$ for readability and write $\mathbf{D}:=\hat{\mathbf{Z}}-\mathbf{z}=\mathbf{Y}+\mathbf{W}$.\par
We first define the event on which $\hat{\mathbf{Z}}$ is close to $\mathbf{z}$, $G:=\left\{|\mathbf{Y}|\leq\delta/2\right\}\cap\left\{|\mathbf{W}|\leq\delta/2\right\}$. Note that whenever $G$ is true, $\left|\mathbf{\hat{Z}}-\mathbf{z}\right|\leq\delta$. By Markov's inequality,
\begin{align}\label{eq:badevent}
    \mathbb{P}\left(G^c\right)&\leq\mathbb{P}\left(|\mathbf{Y}|>\frac{\delta}{2}\right)+\mathbb{P}\left(|\mathbf{W}|>\frac{\delta}{2}\right)=\mathbb{P}\left(|\mathbf{Y}|^2>\frac{\delta^2}{4}\right)+\mathbb{P}\left(|\mathbf{W}|>\frac{\delta}{2}\right)\nonumber\\
    &\leq\frac{4c_1}{\delta^2T}+\frac{2c_2}{\delta T}=:\frac{c_3}{T}.
\end{align}
Using that $G$ implies that the line segment between $\mathbf{\hat{Z}}$ and $\mathbf{z}$ lies in the ball of radius $\delta$ around $\mathbf{z}\in\mathcal{K}$, and hence in $\mathcal{K}_\delta$, the mean value theorem and Taylor's theorem then give, on $G$,
\begin{equation}\label{eq:taylor}
    \left|F(\hat{\mathbf{Z}})-F(\mathbf{z})\right|\leq L_1|\mathbf{D}|,\qquad F(\hat{\mathbf{Z}})-F(\mathbf{z})=\nabla F(\mathbf{z})\cdot\mathbf{D}+R,\qquad |R|\leq\frac{L_2}{2}|\mathbf{D}|^2.
\end{equation}
\begin{enumerate}[(a)]
    
    \item Using \eqref{eq:taylor}, we split the bias into four terms,
    \begin{align}
        &\mathbb{E}\left[F(\hat{\mathbf{Z}})\right]-F(\mathbf{z})=\mathbb{E}\left[\left(F(\hat{\mathbf{Z}})-F(\mathbf{z})\right)\left(\mathbbm{1}_G+\mathbbm{1}_{G^c}\right)\right]\nonumber\\
        &=\mathbb{E}\left[\left(\nabla F(\mathbf{z})\cdot\mathbf{D}+R\right)\mathbbm{1}_G\right]+\mathbb{E}\left[\left(F(\hat{\mathbf{Z}})-F(\mathbf{z})\right)\mathbbm{1}_{G^c}\right]\nonumber\\
        &=\underbrace{\mathbb{E}\left[\left(\nabla F(\mathbf{z})\cdot\mathbf{Y}\right)\mathbbm{1}_G\right]}_{I_1}+\underbrace{\mathbb{E}\left[\left(\nabla F(\mathbf{z})\cdot\mathbf{W}\right)\mathbbm{1}_G\right]}_{I_2}+\underbrace{\mathbb{E}\left[R\,\mathbbm{1}_G\right]}_{I_3}+\underbrace{\mathbb{E}\left[\left(F(\hat{\mathbf{Z}})-F(\mathbf{z})\right)\mathbbm{1}_{G^c}\right]}_{I_4}.
    \end{align}
    Since $\mathbb{E}\left[\mathbf{Y}\right]=0$, we have
    \begin{equation}
        I_1=\mathbb{E}\left[\left(\nabla F(\mathbf{z})\cdot\mathbf{Y}\right)\right]-\mathbb{E}\left[\left(\nabla F(\mathbf{z})\cdot\mathbf{Y}\right)\mathbbm{1}_{G^c}\right]=-\mathbb{E}\left[\left(\nabla F(\mathbf{z})\cdot\mathbf{Y}\right)\mathbbm{1}_{G^c}\right],
    \end{equation}
    so by Cauchy-Schwarz and \eqref{eq:badevent},
    \begin{equation}
        |I_1|\leq \mathbb{E}\left[\left|\nabla F(\mathbf{z}) \right|\left|\mathbf{Y} \right|\mathbbm{1}_{G^c}\right]\leq L_1 \mathbb{E}\left[\left|\mathbf{Y} \right|\mathbbm{1}_{G^c}\right]\leq L_1\sqrt{\mathbb{E}\left[|\mathbf{Y}|^2\right]}\sqrt{\mathbb{P}\left(G^c\right)}\leq\frac{L_1\sqrt{c_1c_3}}{T}.
    \end{equation}
    Next, $|I_2|\leq L_1\mathbb{E}\left[|\mathbf{W}|\right]\leq L_1c_2/T$. On $G$, $|\mathbf{D}|^2\leq2|\mathbf{Y}|^2+2|\mathbf{W}|^2\leq2|\mathbf{Y}|^2+\delta|\mathbf{W}|$, so by \eqref{eq:taylor}
    \begin{equation}
        |I_3|\leq\frac{L_2}{2}\mathbb{E}\left[2|\mathbf{Y}|^2+\delta|\mathbf{W}|\right]\leq\frac{L_2\left(c_1+\delta c_2/2\right)}{T}.
    \end{equation}
    Finally, since $|F(\hat{\mathbf{Z}})|\leq U$ a.s. and $|F(\mathbf{z})|\leq U$, $|I_4|\leq2U\,\mathbb{P}\left(G^c\right)\leq2Uc_3/T$. Summing the four bounds,
    \begin{equation}
        \left|\mathbb{E}\left[F(\hat{\mathbf{Z}})\right]-F(\mathbf{z})\right|\leq\frac{L_1\sqrt{c_1c_3}+L_1c_2+L_2\left(c_1+\delta c_2/2\right)+2Uc_3}{T}.
    \end{equation}
    \item We split the mean squared error on $G$ and $G^c$. On $G$, \eqref{eq:taylor} and $|\mathbf{D}|^2\leq2|\mathbf{Y}|^2+\delta|\mathbf{W}|$ give $\left(F(\hat{\mathbf{Z}})-F(\mathbf{z})\right)^2\leq L_1^2\left(2|\mathbf{Y}|^2+\delta|\mathbf{W}|\right)$. On $G^c$, $\left(F(\hat{\mathbf{Z}})-F(\mathbf{z})\right)^2\leq4U^2$. Hence, using \eqref{eq:badevent},
    \begin{align}
        \mathbb{E}\left[\left(F(\hat{\mathbf{Z}})-F(\mathbf{z})\right)^2\right]&\leq L_1^2\left(2\mathbb{E}\left[|\mathbf{Y}|^2\right]+\delta\,\mathbb{E}\left[|\mathbf{W}|\right]\right)+4U^2\,\mathbb{P}\left(G^c\right)\nonumber\\
        &\leq\frac{L_1^2\left(2c_1+\delta c_2\right)+4U^2c_3}{T}.
    \end{align}
\end{enumerate}
 
\end{proof}
\begin{definition}[population counterparts]\label{def:population-counterparts} For each non-empty subset $A\subseteq[n]$, let $\mathbf{x}^A$ denote either the recorded average $\mathbf{v}^A$ or its noise-free counterpart $\mathbf{v}^{*A}$. We define the population counterparts of the time statistics of $\mathbf{x}^A$ as
\begin{align}
    \overline{\mathrm{Cov}}\left(\mathbf{x}^A,\mathbf{x}^B\right)&:=\widehat{\mathrm{Cov}}_t\left(\mathbf{v}^{*A},\mathbf{v}^{*B}\right)+\begin{cases}
        s(A,B) & \text{if } \mathbf{x}^A=\mathbf{v}^A \text{ and } \mathbf{x}^B=\mathbf{v}^B,\\
        0 & \text{otherwise},
    \end{cases}\label{eq:popcov}\\
    \overline{\langle\mathbf{x}^A\rangle}&:=\langle\mathbf{v}^{*A}\rangle,\label{eq:popmean}
\end{align}
and $\overline{\mathrm{Var}}\left(\mathbf{x}^A\right):=\overline{\mathrm{Cov}}\left(\mathbf{x}^A,\mathbf{x}^A\right)$. These quantities are deterministic.
\end{definition}
 
\begin{theorem}[general limit theorem]\label{thm:general-limit-theorem} Let $\hat{\mathbf{Z}}_T\in\mathbb{R}^d$ collect finitely many statistics, each of which is either of the form $\widehat{\mathrm{Cov}}_t\left(\mathbf{x}^A,\mathbf{x}^B\right)$ or $\langle\mathbf{x}^A\rangle$, or the average $\frac{1}{m}\sum_{i=1}^mS_i$ of an arbitrary number $m$ of statistics $S_1,...,S_m$ of these forms. Let $\mathbf{z}_T$ collect their population counterparts \eqref{eq:popcov}-\eqref{eq:popmean}, where the population counterpart of an average is the average of the population counterparts. Let $\mathcal{K}$, $\delta$, $F$ and $U$ satisfy the conditions of Lemma~\ref{lem:delta-method}. Under Assumptions~\ref{ass:independent-trials}--\ref{ass:summable-fourth-cumulants},
\begin{equation}
    F(\hat{\mathbf{Z}}_T)=F(\mathbf{z}_T)+\varepsilon_T,
\end{equation}
where the constant $c$ of the error term only depends on $M$, $C$, $K_4$, $d$, $\delta$, $U$, $L_1$ and $L_2$, and not on $n$ nor on the number $m$ of averaged statistics.
\end{theorem}
\begin{proof}
By Lemma~\ref{lem:delta-method}, it is sufficient to decompose $\hat{\mathbf{Z}}_T-\mathbf{z}_T=\mathbf{Y}_T+\mathbf{W}_T$ with $\mathbb{E}\left[\mathbf{Y}_T\right]=0$, $\mathbb{E}\left[|\mathbf{Y}_T|^2\right]\leq c_1/T$ and $\mathbb{E}\left[|\mathbf{W}_T|\right]\leq c_2/T$. Since $\mathbb{E}\left[|\mathbf{Y}_T|^2\right]=\sum_{j=1}^d\mathbb{E}\left[Y_j^2\right]$ and $\mathbb{E}\left[|\mathbf{W}_T|\right]\leq\sum_{j=1}^d\mathbb{E}\left[|W_j|\right]$, it is sufficient to do this for each entry $j$ separately. Using $\mathbf{v}^A=\mathbf{v}^{*A}+\boldsymbol{\eta}^A$ and the bilinearity of $\widehat{\mathrm{Cov}}_t$, there are five cases.
\begin{enumerate}[(1)]
    \item \textbf{Time average} $\langle\mathbf{x}^A\rangle$. If $\mathbf{x}^A=\mathbf{v}^{*A}$ the difference with the population counterpart is $0$. If $\mathbf{x}^A=\mathbf{v}^A$ the difference is $\langle\boldsymbol{\eta}^A\rangle$, and we take $Y_j=\langle\boldsymbol{\eta}^A\rangle$, $W_j=0$. By Lemma~\ref{lem:noise-statistics}(a), $\mathbb{E}\left[Y_j\right]=0$ and $\mathbb{E}\left[Y_j^2\right]\leq C/T$.
    \item \textbf{Noise-free covariance} $\widehat{\mathrm{Cov}}_t\left(\mathbf{v}^{*A},\mathbf{v}^{*B}\right)$. The difference is $0$.
    \item \textbf{Mixed covariance} $\widehat{\mathrm{Cov}}_t\left(\mathbf{v}^{*A},\mathbf{v}^{B}\right)$ (or symmetrically). The difference is $\widehat{\mathrm{Cov}}_t\left(\mathbf{v}^{*A},\boldsymbol{\eta}^{B}\right)$, and we take it as $Y_j$, with $W_j=0$. By Lemma~\ref{lem:noise-statistics}(b), $\mathbb{E}\left[Y_j\right]=0$ and $\mathbb{E}\left[Y_j^2\right]\leq4M^2C/T$.
    \item \textbf{Recorded covariance} $\widehat{\mathrm{Cov}}_t\left(\mathbf{v}^{A},\mathbf{v}^{B}\right)$. The difference is
    \begin{equation}
        \widehat{\mathrm{Cov}}_t\left(\mathbf{v}^{*A},\boldsymbol{\eta}^{B}\right)+\widehat{\mathrm{Cov}}_t\left(\boldsymbol{\eta}^{A},\mathbf{v}^{*B}\right)+Y_{AB}+W_{AB},
    \end{equation}
    with $Y_{AB}$ and $W_{AB}$ defined in Lemma~\ref{lem:noise-statistics}(c). We take $Y_j$ to be the sum of the first three terms and $W_j=W_{AB}$. By Lemma~\ref{lem:noise-statistics}(b)-(c), $\mathbb{E}\left[Y_j\right]=0$, and by Minkowski's inequality
    \begin{align}
        \sqrt{\mathbb{E}\left[Y_j^2\right]}&\leq\sqrt{\mathbb{E}\left[\widehat{\mathrm{Cov}}_t\left(\mathbf{v}^{*A},\boldsymbol{\eta}^{B}\right)^2\right]}+\sqrt{\mathbb{E}\left[\widehat{\mathrm{Cov}}_t\left(\boldsymbol{\eta}^{A},\mathbf{v}^{*B}\right)^2\right]}+\sqrt{\mathbb{E}\left[Y_{AB}^2\right]}\nonumber\\
        &\leq2\sqrt{\frac{4M^2C}{T}}+\sqrt{\frac{C^2+K_4}{T}}\nonumber\\
        &=\frac{4M\sqrt{C}+\sqrt{C^2+K_4}}{\sqrt{T}},
    \end{align}
    while $\mathbb{E}\left[|W_j|\right]\leq C/T$.
    \item \textbf{Average} $\frac{1}{m}\sum_{i=1}^mS_i$ of statistics of the forms (1)-(4). Each difference between $S_i$ and its population counterpart decomposes as $Y^{(i)}+W^{(i)}$ as above. We take $Y_j:=\frac{1}{m}\sum_iY^{(i)}$ and $W_j:=\frac{1}{m}\sum_iW^{(i)}$. Then $\mathbb{E}\left[Y_j\right]=0$, and by Minkowski's inequality and the triangle inequality,
    \begin{equation}
        \sqrt{\mathbb{E}\left[Y_j^2\right]}\leq\frac{1}{m}\sum_{i=1}^m\sqrt{\mathbb{E}\left[\left(Y^{(i)}\right)^2\right]},\qquad\mathbb{E}\left[|W_j|\right]\leq\frac{1}{m}\sum_{i=1}^m\mathbb{E}\left[\left|W^{(i)}\right|\right],
    \end{equation}
    so that the bounds of cases (1)-(4) also hold for $Y_j$ and $W_j$, independently of $m$.
\end{enumerate}
Hence the conditions of Lemma~\ref{lem:delta-method} hold with $c_1=d\left[\left(4M\sqrt{C}+\sqrt{C^2+K_4}\right)^2+C\right]$ and $c_2=dC$, which do not depend on $n$, and Lemma~\ref{lem:delta-method} gives the result.
 
\end{proof}
\begin{corollary}[correlation of trial averages]\label{cor:correlation-of-trial-averages} Let $A,B\subseteq[n]$ be non-empty and assume that there exists a constant $a_0>0$ such that
\begin{equation}\label{eq:nondeg}
    \widehat{\mathrm{Var}}_t\left(\mathbf{v}^{*A}\right)\geq a_0,\qquad\widehat{\mathrm{Var}}_t\left(\mathbf{v}^{*B}\right)\geq a_0\qquad\text{for all } T.
\end{equation}
Under Assumptions~\ref{ass:independent-trials}--\ref{ass:summable-fourth-cumulants},
\begin{align}
    \widehat{\mathrm{Corr}}_t\left(\mathbf{v}^{A},\mathbf{v}^{B}\right)&=\frac{\widehat{\mathrm{Cov}}_t\left(\mathbf{v}^{*A},\mathbf{v}^{*B}\right)+s(A,B)}{\sqrt{\left(\widehat{\mathrm{Var}}_t\left(\mathbf{v}^{*A}\right)+s(A,A)\right)\left(\widehat{\mathrm{Var}}_t\left(\mathbf{v}^{*B}\right)+s(B,B)\right)}}+\varepsilon_T,\label{eq:corr-rec}\\
    \widehat{\mathrm{Corr}}_t\left(\mathbf{v}^{*A},\mathbf{v}^{B}\right)&=\frac{\widehat{\mathrm{Cov}}_t\left(\mathbf{v}^{*A},\mathbf{v}^{*B}\right)}{\sqrt{\widehat{\mathrm{Var}}_t\left(\mathbf{v}^{*A}\right)\left(\widehat{\mathrm{Var}}_t\left(\mathbf{v}^{*B}\right)+s(B,B)\right)}}+\varepsilon_T,\label{eq:corr-free}
\end{align}
where in both cases the constant $c$ of the error term only depends on $M$, $C$, $K_4$ and $a_0$.
\end{corollary}
\begin{proof}
We prove \eqref{eq:corr-rec}; the proof of \eqref{eq:corr-free} is identical with $s(A,B)$ and $s(A,A)$ set to zero. We apply Theorem~\ref{thm:general-limit-theorem} with $d=3$,
\begin{equation}
    \hat{\mathbf{Z}}_T=\left(\widehat{\mathrm{Cov}}_t\left(\mathbf{v}^{A},\mathbf{v}^{B}\right),\widehat{\mathrm{Var}}_t\left(\mathbf{v}^{A}\right),\widehat{\mathrm{Var}}_t\left(\mathbf{v}^{B}\right)\right),
\end{equation}
\begin{equation}
    F(u_1,u_2,u_3):=\begin{cases}
        \frac{u_1}{\sqrt{u_2u_3}} & \text{if } u_2>0 \text{ and } u_3>0,\\
        0 & \text{otherwise},
    \end{cases}
\end{equation}
so that $F(\hat{\mathbf{Z}}_T)=\widehat{\mathrm{Corr}}_t\left(\mathbf{v}^{A},\mathbf{v}^{B}\right)$ (with the convention that the correlation is $0$ if one of the variances vanishes). The population counterpart is
\begin{align}
    \mathbf{z}_T=\Big(&\widehat{\mathrm{Cov}}_t\left(\mathbf{v}^{*A},\mathbf{v}^{*B}\right)+s(A,B),\nonumber\\
    &\widehat{\mathrm{Var}}_t\left(\mathbf{v}^{*A}\right)+s(A,A),\ \widehat{\mathrm{Var}}_t\left(\mathbf{v}^{*B}\right)+s(B,B)\Big),
\end{align}
and $F(\mathbf{z}_T)$ is the leading term of \eqref{eq:corr-rec}. We check the conditions of Lemma~\ref{lem:delta-method}:
\begin{enumerate}[(a)]
    \item \textbf{Compact set} By Assumption~\ref{ass:bounded-signals}, $\widehat{\mathrm{Var}}_t\left(\mathbf{v}^{*A}\right)=\langle\left(\mathbf{v}^{*A}\right)^2\rangle-\langle\mathbf{v}^{*A}\rangle^2\leq\langle\left(\mathbf{v}^{*A}\right)^2\rangle\leq M^2$, and similarly for $B$, so that by Cauchy-Schwarz $\left|\widehat{\mathrm{Cov}}_t\left(\mathbf{v}^{*A},\mathbf{v}^{*B}\right)\right|\leq M^2$. Moreover, $\sigma_\ell^2\leq C$ (it is an average of the $\gamma_\ell(t,t)\leq C$), hence $0\leq s(A,B)\leq\frac{|A\cap B|}{|A||B|}C\leq C$. Together with \eqref{eq:nondeg}, $\mathbf{z}_T$ lies in the compact set
    \begin{equation}
        \mathcal{K}:=\left[-M^2-C,\,M^2+C\right]\times\left[a_0,\,M^2+C\right]\times\left[a_0,\,M^2+C\right].
    \end{equation}
    \item \textbf{Smoothness} We take $\delta:=a_0/2$. Every point $(u_1,u_2,u_3)$ of $\mathcal{K}_\delta$ satisfies $u_2\geq a_0/2$ and $u_3\geq a_0/2$, and $F$ is infinitely differentiable on the open set $\left\{(u_1,u_2,u_3): u_2>a_0/4,\ u_3>a_0/4\right\}$, which contains $\mathcal{K}_\delta$.
    \item \textbf{Boundedness} $|F(\hat{\mathbf{Z}}_T)|\leq1$ since it is a Pearson correlation. For the population counterpart, let $\tau$ be uniformly distributed on $[T]$ and let $\xi_1,\dots,\xi_n$ be independent, centered random variables with $\mathrm{Var}(\xi_\ell)=\sigma_\ell^2$, independent of $\tau$. With $\xi^A:=\frac{1}{|A|}\sum_{\ell\in A}\xi_\ell$ (similarly for $\xi^B$), set $X:=v^{*A}_\tau+\xi^A$ and $Y:=v^{*B}_\tau+\xi^B$. Then $F(\mathbf{z}_T)=\mathrm{Corr}(X,Y)$, so $|F(\mathbf{z}_T)|\leq1$.
\end{enumerate}
All conditions of Lemma~\ref{lem:delta-method} hold, and Theorem~\ref{thm:general-limit-theorem} gives the result.
 
\end{proof}
\begin{lemma}[thresholded functions]\label{lem:thresholded-functions} Let $X_T=x_T+\varepsilon_T$ with $x_T$ deterministic and $\varepsilon_T$ with $\mathbb{E}\left[\varepsilon_T^2\right]\leq c/T$. Then
\begin{equation}
    \sqrt{\mathrm{max}\left(0,X_T\right)}=\sqrt{\mathrm{max}\left(0,x_T\right)}+\varepsilon_T^{\mathrm{max}},
\end{equation}
where the constant of $\varepsilon_T^{\mathrm{max}}$ is the constant $c$ of the assumption.
\end{lemma}
\begin{proof} Let $f(x):=\sqrt{\mathrm{max}(0,x)}$, we first prove that $\left|f(x)-f(y)\right|\leq\sqrt{|x-y|}$. If $x,y\leq 0$, the difference is $0$; if $y\leq0<x$, $|f(x)-f(y)|=f(x)=\sqrt{x}\leq\sqrt{x-y}$ (and symmetrically if $x\leq0<y$); if $x,y>0$, $\left|f(x)-f(y)\right|=|\sqrt{x}-\sqrt{y}|=\sqrt{(\sqrt{x}-\sqrt{y})^2}\leq\sqrt{x+y-2\mathrm{min}(x,y)}=\sqrt{|x-y|}$. This gives
\begin{equation}
    |\varepsilon_T^\mathrm{max}|=|f(X_T)-f(x_T)|\leq\sqrt{|\varepsilon_T|}.
\end{equation}
Hence $(\varepsilon_T^{\mathrm{max}})^2\leq|\varepsilon_T|$, so that $\mathbb{E}[(\varepsilon_T^{\mathrm{max}})^2]\leq\mathbb{E}[|\varepsilon_T|]\leq\sqrt{\mathbb{E}[\varepsilon_T^2]}\leq(c/T)^{1/2}$. Furthermore, by Jensen's inequality
\begin{equation}
    \left|\mathbb{E}\left[\varepsilon_T^{\mathrm{max}}\right]\right|\leq\mathbb{E}\left[|\varepsilon_T^{\mathrm{max}}|\right]\leq\mathbb{E}\left[\sqrt{|\varepsilon_T|}\right]\leq\sqrt{\mathbb{E}\left[|\varepsilon_T|\right]}\leq\left(\mathbb{E}\left[\varepsilon_T^2\right]\right)^{1/4}\leq (c/T)^{1/4}.
\end{equation}
\end{proof}
\begin{remark}[on Lemma~\ref{lem:thresholded-functions}] The slower convergence in Lemma~\ref{lem:thresholded-functions} compared to Lemma~\ref{lem:delta-method} is due to the max function not being differentiable in $0$. In the case where the sequence $\{x_T\}_{T>0}$ is bounded away from zero, one falls back into the setting of Lemma~\ref{lem:delta-method} and the convergence is faster.
\end{remark}
 
\subsection{Latent representation of the noise-free signals}
 
\begin{definition}[latent signals]\label{def:latent-signals} Let $\boldsymbol{\phi}_1,...,\boldsymbol{\phi}_K$ be sequences of length $T$ that are centered and orthonormal over time,
\begin{equation}\label{eq:orthonormal}
    \langle\boldsymbol{\phi}_k\rangle=0,\qquad \langle\boldsymbol{\phi}_j\boldsymbol{\phi}_k\rangle=\begin{cases}
        1 & \text{if } j=k,\\
        0 & \text{otherwise},
    \end{cases}
\end{equation}
and assume that the centered noise-free signal of every trial can be written as
\begin{equation}\label{eq:latent}
    \mathbf{v}^{*(\ell)}-\langle\mathbf{v}^{*(\ell)}\rangle=\sum_{k=1}^Kw_k^{(\ell)}\boldsymbol{\phi}_k,\qquad w_k^{(\ell)}:=\widehat{\mathrm{Cov}}_t\left(\mathbf{v}^{*(\ell)},\boldsymbol{\phi}_k\right).
\end{equation}
For a non-empty subset $A\subseteq[n]$, we write $\bar{w}_k^A:=\frac{1}{|A|}\sum_{\ell\in A}w_k^{(\ell)}$ for the average weights, and $\bar{w}_k:=\bar{w}_k^{[n]}$.\par
Such a representation always exists: the centered noise-free signals $\mathbf{v}^{*(\ell)}-\langle\mathbf{v}^{*(\ell)}\rangle$, $\ell\in[n]$, span a subspace of $\mathbb{R}^T$ of dimension at most $n$, and applying the Gram-Schmidt procedure to them with the inner product $\langle\mathbf{xy}\rangle$ gives centered orthonormal sequences satisfying \eqref{eq:latent}.
\end{definition}
 
\begin{lemma}[latent form of the noise-free covariances]\label{lem:latent-form-of-the} Under the representation of Definition~\ref{def:latent-signals}, for any non-empty $A,B\subseteq[n]$,
\begin{equation}\label{eq:latentcov}
    \widehat{\mathrm{Cov}}_t\left(\mathbf{v}^{*A},\mathbf{v}^{*B}\right)=\sum_{k=1}^K\bar{w}_k^A\bar{w}_k^B,\qquad\text{in particular}\qquad \widehat{\mathrm{Var}}_t\left(\mathbf{v}^{*A}\right)=\sum_{k=1}^K\left(\bar{w}_k^A\right)^2.
\end{equation}
\end{lemma}
\begin{proof}
Averaging \eqref{eq:latent} over $\ell\in A$ gives $\mathbf{v}^{*A}-\langle\mathbf{v}^{*A}\rangle=\sum_k\bar{w}_k^A\boldsymbol{\phi}_k$, and similarly for $B$. Since $\widehat{\mathrm{Cov}}_t$ is bilinear and invariant under the addition of constants,
\begin{equation}
    \widehat{\mathrm{Cov}}_t\left(\mathbf{v}^{*A},\mathbf{v}^{*B}\right)=\widehat{\mathrm{Cov}}_t\left(\sum_{j=1}^K\bar{w}_j^A\boldsymbol{\phi}_j,\sum_{k=1}^K\bar{w}_k^B\boldsymbol{\phi}_k\right)=\sum_{j=1}^K\sum_{k=1}^K\bar{w}_j^A\bar{w}_k^B\,\widehat{\mathrm{Cov}}_t\left(\boldsymbol{\phi}_j,\boldsymbol{\phi}_k\right)=\sum_{k=1}^K\bar{w}_k^A\bar{w}_k^B,
\end{equation}
where the last equality uses \eqref{eq:orthonormal}, which gives $\widehat{\mathrm{Cov}}_t\left(\boldsymbol{\phi}_j,\boldsymbol{\phi}_k\right)=\langle\boldsymbol{\phi}_j\boldsymbol{\phi}_k\rangle-\langle\boldsymbol{\phi}_j\rangle\langle\boldsymbol{\phi}_k\rangle=\langle\boldsymbol{\phi}_j\boldsymbol{\phi}_k\rangle$.
 
\end{proof}
\subsection{Trial-averaged noise ceiling}
In this section, we derive the closed-form expressions for the true noise ceiling for trial-averaged responses $NC^{\mathrm{mean}}$ and its estimators $\widehat{NC}^{\mathrm{SHSB}}$ and $\widehat{NC}^{\mathrm{ATV}}$.
\subsubsection{True noise ceiling ${NC}^{\mathrm{mean}}$}\label{sec:true_noise_ceiling}
We recall that the true noise ceiling for trial-averaged responses is the correlation between the noise-free average and the recorded average, $NC^{\mathrm{mean}}:=\widehat{\mathrm{Corr}}_t\left(\bar{\mathbf{v}}^*,\bar{\mathbf{v}}\right)$, with $\bar{\mathbf{v}}^*:=\mathbf{v}^{*[n]}$ and $\bar{\mathbf{v}}:=\mathbf{v}^{[n]}$. We write
\begin{equation}
    \bar{\sigma}^2:=\frac{1}{n}\sum_{\ell=1}^n\sigma_\ell^2
\end{equation}
for the noise variance averaged over trials and time.
 
\begin{proposition}[true noise ceiling, trial-averaged responses]\label{prop:true-noise-ceiling-trial-averaged} Assume that there exists a constant $a_0>0$ such that $\widehat{\mathrm{Var}}_t\left(\bar{\mathbf{v}}^*\right)\geq a_0$ for all $T$. Under Assumptions~\ref{ass:independent-trials}--\ref{ass:summable-fourth-cumulants},
\begin{equation}\label{eq:NCmean}
    NC^{\mathrm{mean}}=\sqrt{\frac{\widehat{\mathrm{Var}}_t\left(\bar{\mathbf{v}}^*\right)}{\widehat{\mathrm{Var}}_t\left(\bar{\mathbf{v}}^*\right)+\bar{\sigma}^2/n}}+\varepsilon_T=\sqrt{\frac{\sum_{k=1}^K\bar{w}_k^2}{\sum_{k=1}^K\bar{w}_k^2+\bar{\sigma}^2/n}}+\varepsilon_T,
\end{equation}
where the constant $c$ of the error term only depends on $M$, $C$, $K_4$ and $a_0$, and not on $n$. The second expression holds under the representation of Definition~\ref{def:latent-signals}.
\end{proposition}
\begin{proof}
We apply \eqref{eq:corr-free} of Corollary~\ref{cor:correlation-of-trial-averages} with $A=B=[n]$. The shared noise of the full average with itself is
\begin{equation}
    s([n],[n])=\frac{1}{n^2}\sum_{\ell=1}^n\sigma_\ell^2=\frac{\bar{\sigma}^2}{n},
\end{equation}
and $\widehat{\mathrm{Cov}}_t\left(\bar{\mathbf{v}}^*,\bar{\mathbf{v}}^*\right)=\widehat{\mathrm{Var}}_t\left(\bar{\mathbf{v}}^*\right)$. Corollary~\ref{cor:correlation-of-trial-averages} therefore gives
\begin{equation}
    NC^{\mathrm{mean}}=\frac{\widehat{\mathrm{Var}}_t\left(\bar{\mathbf{v}}^*\right)}{\sqrt{\widehat{\mathrm{Var}}_t\left(\bar{\mathbf{v}}^*\right)\left(\widehat{\mathrm{Var}}_t\left(\bar{\mathbf{v}}^*\right)+\bar{\sigma}^2/n\right)}}+\varepsilon_T=\sqrt{\frac{\widehat{\mathrm{Var}}_t\left(\bar{\mathbf{v}}^*\right)}{\widehat{\mathrm{Var}}_t\left(\bar{\mathbf{v}}^*\right)+\bar{\sigma}^2/n}}+\varepsilon_T,
\end{equation}
which is the first expression in \eqref{eq:NCmean}. The second expression follows from Lemma~\ref{lem:latent-form-of-the} with $A=[n]$, $\widehat{\mathrm{Var}}_t\left(\bar{\mathbf{v}}^*\right)=\sum_k\bar{w}_k^2$.
 
\end{proof}

\subsubsection{Across-trial variance method $\widehat{NC}^{\mathrm{ATV}}$}\label{sec:ATV}
We assume that each trial is centered over time before the estimator is computed, i.e. $\mathbf{v}^{(\ell)}$ is replaced by $\mathbf{v}^{(\ell)}-\langle\mathbf{v}^{(\ell)}\rangle$. This is the centering step of the usual z-scoring of each run or subject; the division by the standard deviation is not needed. With the unbiased variance across trials at each time step,
\begin{equation}
    \widehat{\mathrm{Var}}^{\mathrm{u}}_\ell\left[v_t\right]:=\frac{1}{n-1}\sum_{\ell=1}^n\left(v_t^{(\ell)}-\bar{v}_t\right)^2,
\end{equation}
the ATV estimator of Table~\ref{tab:nc_methods} is
\begin{equation}\label{eq:ATV}
    \widehat{NC}^{\mathrm{ATV}}:=\sqrt{\mathrm{max}\left(0,\hat{F}\right)},\qquad\hat{F}:=1-\frac{1}{n}\cdot\frac{\frac{1}{T}\sum_{t=1}^T\widehat{\mathrm{Var}}^{\mathrm{u}}_\ell\left[v_t\right]}{\widehat{\mathrm{Var}}_t\left(\bar{\mathbf{v}}\right)},
\end{equation}
with the convention $\widehat{NC}^{\mathrm{ATV}}=0$ if $\widehat{\mathrm{Var}}_t\left(\bar{\mathbf{v}}\right)=0$. For a quantity $x^{(\ell)}$ defined for each trial, we write $\widehat{\mathrm{Var}}_\ell\left[x\right]:=\frac{1}{n}\sum_{\ell=1}^n\left(x^{(\ell)}-\frac{1}{n}\sum_{\ell'=1}^nx^{(\ell')}\right)^2$ for its empirical variance across trials.
 
\begin{proposition}[ATV]\label{prop:ATV} Assume that there exists a constant $a_0>0$ such that $\widehat{\mathrm{Var}}_t\left(\bar{\mathbf{v}}^*\right)\geq a_0$ for all $T$. Under Assumptions~\ref{ass:independent-trials}--\ref{ass:summable-fourth-cumulants}:
\begin{enumerate}[(a)]
    \item \textbf{(general case)}
    \begin{equation}
        \widehat{NC}^{\mathrm{ATV}}=\sqrt{\mathrm{max}\left(0,\frac{\sum_k\bar{w}_k^2-\frac{1}{n-1}\sum_k\widehat{\mathrm{Var}}_\ell\left[w_k\right]}{\sum_k\bar{w}_k^2+\bar{\sigma}^2/n}\right)}+\varepsilon_T^{\mathrm{max}};
    \end{equation}
    \item \textbf{(units with signal)} if there exists $f_{\mathrm{min}}>0$ such that $\frac{\sum_k\bar{w}_k^2-\frac{1}{n-1}\sum_k\widehat{\mathrm{Var}}_\ell\left[w_k\right]}{\sum_k\bar{w}_k^2+\bar{\sigma}^2/n}\geq f_{\mathrm{min}}$ for all $T$, then
    \begin{equation}\label{eq:ATV-result}
        \widehat{NC}^{\mathrm{ATV}}=\sqrt{\frac{\sum_k\bar{w}_k^2-\frac{1}{n-1}\sum_k\widehat{\mathrm{Var}}_\ell\left[w_k\right]}{\sum_k\bar{w}_k^2+\bar{\sigma}^2/n}}+\varepsilon_T.
    \end{equation}
\end{enumerate}
The constant $c$ of the error terms only depends on $M$, $C$, $K_4$, $a_0$ (and $f_{\mathrm{min}}$ in (b)), and not on $n$.\par

\end{proposition}
\begin{proof}
\begin{enumerate}[(1)]
    \item \textbf{Rewriting the estimator} After centering, $\langle\mathbf{v}^{(\ell)}\rangle=0$ for all $\ell$ and $\langle\bar{\mathbf{v}}\rangle=0$, so time averages of squares are variances over time. Since $\sum_\ell\left(v_t^{(\ell)}-\bar{v}_t\right)^2=\sum_\ell\left(v_t^{(\ell)}\right)^2-n\bar{v}_t^2$,
    \begin{equation}
        \frac{1}{T}\sum_{t=1}^T\widehat{\mathrm{Var}}^{\mathrm{u}}_\ell\left[v_t\right]=\frac{1}{n-1}\left(\sum_{\ell=1}^n\widehat{\mathrm{Var}}_t\left(\mathbf{v}^{(\ell)}\right)-n\,\widehat{\mathrm{Var}}_t\left(\bar{\mathbf{v}}\right)\right).
    \end{equation}
    Hence $\hat{F}=F\left(\hat{\mathbf{Z}}_T\right)$ with
    \begin{equation}
        \hat{\mathbf{Z}}_T:=\left(\widehat{\mathrm{Var}}_t\left(\bar{\mathbf{v}}\right),\ \frac{1}{n}\sum_{\ell=1}^n\widehat{\mathrm{Var}}_t\left(\mathbf{v}^{(\ell)}\right)\right),\qquad F(u_1,u_2):=1-\frac{u_2-u_1}{(n-1)u_1}.
    \end{equation}
    Using Cauchy-Schwarz, the second entry of $\hat{\mathbf{Z}}_T$ is at least the first one, which gives $\hat{F}\leq1$.
    \item \textbf{Population counterparts} The first entry of $\hat{\mathbf{Z}}_T$ is a statistic of the form $\widehat{\mathrm{Cov}}_t\left(\mathbf{x}^A,\mathbf{x}^B\right)$ and the second one is the average of the $n$ statistics $\widehat{\mathrm{Var}}_t\left(\mathbf{v}^{(\ell)}\right)$. By Definition~\ref{def:population-counterparts} and Lemma~\ref{lem:latent-form-of-the}, their population counterparts are
    \begin{align}
        \mathbf{z}_T&:=\left(\sum_k\bar{w}_k^2+\frac{\bar{\sigma}^2}{n},\ \frac{1}{n}\sum_{\ell=1}^n\left(\sum_k\left(w_k^{(\ell)}\right)^2+\sigma_\ell^2\right)\right)\nonumber\\
        &=\left(\sum_k\bar{w}_k^2+\frac{\bar{\sigma}^2}{n},\ \sum_k\bar{w}_k^2+\sum_k\widehat{\mathrm{Var}}_\ell\left[w_k\right]+\bar{\sigma}^2\right),
    \end{align}
    where we used $\frac{1}{n}\sum_\ell\left(w_k^{(\ell)}\right)^2=\bar{w}_k^2+\widehat{\mathrm{Var}}_\ell\left[w_k\right]$. Therefore
    \begin{equation}
        F\left(\mathbf{z}_T\right)=1-\frac{\frac{1}{n-1}\sum_k\widehat{\mathrm{Var}}_\ell\left[w_k\right]+\bar{\sigma}^2/n}{\sum_k\bar{w}_k^2+\bar{\sigma}^2/n}=\frac{\sum_k\bar{w}_k^2-\frac{1}{n-1}\sum_k\widehat{\mathrm{Var}}_\ell\left[w_k\right]}{\sum_k\bar{w}_k^2+\bar{\sigma}^2/n}.
    \end{equation}
    \item \textbf{Compact set} Using Assumption~\ref{ass:bounded-signals}, $\sum_k \bar{w}_k^2\leq\frac{1}{n}\sum_\ell\sum_kw_k^{(\ell)2}\leq\max_\ell\sum_kw_k^{(\ell)2}=\widehat{\mathrm{Var}}_t\left(\mathbf{v}^{*(\ell)}\right)\leq M^2$. Moreover by Assumption~\ref{ass:short-range-temporal-correlations}, $\bar{\sigma}^2\leq C$. Therefore $\mathbf{z}_T$ lies in $\mathcal{K}_0:=\left[a_0,M^2+C\right]\times\left[a_0,M^2+C\right]$. Additionally, since $\frac{1}{n-1}\sum_k\widehat{\mathrm{Var}}_\ell\left[w_k\right]\leq\frac{1}{n}\sum_\ell\sum_k\left(w_k^{(\ell)}\right)^2\leq M^2$, $F\left(\mathbf{z}_T\right)\geq-M^2/a_0$.
    \item \textbf{Smoothness} We write $F=1-\frac{1}{n-1}h$ with $h(u_1,u_2):=\frac{u_2-u_1}{u_1}$, which does not depend on $n$. $h$ is infinitely differentiable on the open set $\left\{(u_1,u_2):u_1>a_0/4\right\}$, which contains the compact set $(\mathcal{K}_0)_{a_0/2}$. Hence $h$ is uniformly continuous on $(\mathcal{K}_0)_{a_0/2}$ (since $(\mathcal{K}_0)_{a_0/2}$ is compact), and $F$ is infinitely differentiable on any $(\mathcal{K}_0)_\delta\subseteq(\mathcal{K}_0)_{a_0/2}$ with first and second derivatives bounded independently of $n\geq2$, since $\frac{1}{n-1}\leq1$.
    \item \textbf{Proof of (b)} Let $\mathcal{K}:=\left\{\mathbf{u}\in\mathcal{K}_0:F(\mathbf{u})\geq f_{\mathrm{min}}\right\}$, which is compact and contains all $\mathbf{z}_T$, and choose $\delta\leq a_0/2$ such that $|h(\mathbf{x})-h(\mathbf{y})|\leq f_{\mathrm{min}}/2$ whenever $\mathbf{x},\mathbf{y}\in\left(\mathcal{K}_0\right)_{a_0/2}$ and $|\mathbf{x}-\mathbf{y}|\leq\delta$ (possible because $h$ is uniformly continuous). Since $\frac{1}{n-1}\leq1$, $\delta$ does not depend on $n$, and $F\geq f_{\mathrm{min}}/2$ on $\mathcal{K}_\delta$, so that $G:=\sqrt{\mathrm{max}(0,F)}=\sqrt{F}$ is infinitely differentiable on an open set containing $\mathcal{K}_\delta$. Moreover $0\leq G\left(\hat{\mathbf{Z}}_T\right)\leq1$ since $\hat{F}\leq1$, and $0\leq G\left(\mathbf{z}_T\right)\leq1$. Theorem~\ref{thm:general-limit-theorem} applied to $G$ gives (b).
    \item \textbf{Proof of (a)} $F$ is not bounded near $u_1=0$, so we first clip it: let $\tilde{F}:=\mathrm{max}\left(F,-1-M^2/a_0\right)$. Let $\mathcal{K}':=\left\{\mathbf{u}\in\mathcal{K}_0:F(\mathbf{u})\geq-M^2/a_0\right\}$, which is compact and contains all $\mathbf{z}_T$ by (3), and choose $\delta\leq a_0/2$ as in (5), with $1/2$ instead of $f_{\mathrm{min}}/2$. On $\mathcal{K}'_\delta$, $F\geq-M^2/a_0-1/2$, so $\tilde{F}=F$ on an open set containing $\mathcal{K}'_\delta$, where it is infinitely differentiable. Moreover $\left|\tilde{F}\left(\hat{\mathbf{Z}}_T\right)\right|\leq1+M^2/a_0$ and $\left|\tilde{F}\left(\mathbf{z}_T\right)\right|\leq1+M^2/a_0$. Theorem~\ref{thm:general-limit-theorem} applied to $\tilde{F}$ gives $\mathbb{E}\left[\left(\tilde{F}\left(\hat{\mathbf{Z}}_T\right)-F\left(\mathbf{z}_T\right)\right)^2\right]\leq c/T$. Since $\tilde{F}$ and $F$ coincide where they are positive, $\widehat{NC}^{\mathrm{ATV}}=\sqrt{\mathrm{max}\left(0,\tilde{F}\left(\hat{\mathbf{Z}}_T\right)\right)}$, and Lemma~\ref{lem:thresholded-functions} with $X_T=\tilde{F}\left(\hat{\mathbf{Z}}_T\right)$ and $x_T=F\left(\mathbf{z}_T\right)$ gives (a).
\end{enumerate}
 
\end{proof}
\begin{remark} \textbf{Normalization of the variance across trials} If the variance across trials is computed with normalization $1/n$ instead of $1/(n-1)$, the same computation gives $\frac{\sum_k\bar{w}_k^2-\frac{1}{n}\sum_k\widehat{\mathrm{Var}}_\ell\left[w_k\right]+\bar{\sigma}^2/n^2}{\sum_k\bar{w}_k^2+\bar{\sigma}^2/n}$ under the square root. Therefore it results in an additional term $\frac{\frac{1}{n(n-1)}\sum_k\widehat{\mathrm{Var}}_\ell\left[w_k\right]+\bar{\sigma}^2/n^2}{\sum_k\bar{w}_k^2+\bar{\sigma}^2/n}$.

\end{remark}
\subsubsection{Split-half correlation with Spearman-Brown correction $\widehat{NC}^{\mathrm{SHSB}}$}\label{SHSB}
In this section we assume that $n\geq2$ is even. We call a \emph{split} a pair $(A,B)$ of disjoint subsets with $A\cup B=[n]$ and $|A|=|B|=n/2$. For a split, we write
\begin{equation}
    \bar{\sigma}^2_A:=\frac{2}{n}\sum_{\ell\in A}\sigma_\ell^2
\end{equation}
for the noise level of $A$ (similarly for $B$), so that $\frac{1}{2}\left(\bar{\sigma}^2_A+\bar{\sigma}^2_B\right)=\bar{\sigma}^2$, and $\hat{\rho}_{A,B}:=\widehat{\mathrm{Corr}}_t\left(\mathbf{v}^A,\mathbf{v}^B\right)$ for the correlation between the two half averages. The SHSB estimator of Table~\ref{tab:nc_methods} for a single split is
\begin{equation}\label{eq:SHSB}
    \widehat{NC}^{\mathrm{SHSB}}_{A,B}:=\sqrt{\mathrm{max}\left(0,\frac{2\hat{\rho}_{A,B}}{1+\hat{\rho}_{A,B}}\right)}.
\end{equation}
In practice, the estimator is usually computed for a collection $\mathcal{S}$ of $r$ splits (all possible splits or $r$ randomly selected ones) and averaged,
\begin{equation}\label{eq:SHSB-SB}
    \widehat{NC}^{\mathrm{SHSB}}:=\sqrt{\mathrm{max}\left(0,\frac{1}{r}\sum_{(A,B)\in\mathcal{S}}\frac{2\hat{\rho}_{A,B}}{1+\hat{\rho}_{A,B}}\right)}.
\end{equation}
 
\begin{proposition}[SHSB]\label{prop:shsb} Let $(A,B)$ be a split and assume that there exists a constant $a_0>0$ such that
\begin{equation}\label{eq:shsb-cond}
    \widehat{\mathrm{Var}}_t\left(\mathbf{v}^{*A}\right)\geq a_0,\qquad\widehat{\mathrm{Var}}_t\left(\mathbf{v}^{*B}\right)\geq a_0,\qquad\widehat{\mathrm{Cov}}_t\left(\mathbf{v}^{*A},\mathbf{v}^{*B}\right)\geq0\qquad\text{for all } T.
\end{equation}
Under Assumptions~\ref{ass:independent-trials}--\ref{ass:summable-fourth-cumulants}:
\begin{enumerate}[(a)]
    \item \textbf{(general case)}
    \begin{equation}\label{eq:SHSB-result}
        \widehat{NC}^{\mathrm{SHSB}}_{A,B}=\sqrt{\frac{\sum_k\bar{w}_k^2-\Delta_{AB}^{w}}{\sum_k\bar{w}_k^2+\bar{\sigma}^2/n-\Delta_{AB}^{\mathrm{sd}}}}+\varepsilon_T^{\mathrm{max}},
    \end{equation}
    where we used
    \begin{align}
       \Delta_{AB}^{w}&:=\frac{1}{4}\sum_k\left(\bar{w}_k^A-\bar{w}_k^B\right)^2, \nonumber\\\Delta_{AB}^{\mathrm{sd}}&:=\frac{1}{4}\left(\sqrt{\sum_k\left(\bar{w}_k^A\right)^2+\frac{2\bar{\sigma}^2_A}{n}}-\sqrt{\sum_k\left(\bar{w}_k^B\right)^2+\frac{2\bar{\sigma}^2_B}{n}}\right)^2;
    \end{align}
    \item \textbf{(units with signal)} if there exists $f_{\mathrm{min}}>0$ such that the argument of the square root in \eqref{eq:SHSB-result} is at least $f_{\mathrm{min}}$ for all $T$, then \eqref{eq:SHSB-result} holds with $\varepsilon_T$ instead of $\varepsilon_T^{\mathrm{max}}$;
    \item \textbf{(averages over splits)} if \eqref{eq:shsb-cond} holds for every split of $\mathcal{S}$,
    \begin{equation}\label{eq:SHSB-avg-result}
        \widehat{NC}^{\mathrm{SHSB}}=\sqrt{\frac{1}{r}\sum_{(A,B)\in\mathcal{S}}\frac{\sum_k\bar{w}_k^2-\Delta_{AB}^{w}}{\sum_k\bar{w}_k^2+\bar{\sigma}^2/n-\Delta_{AB}^{\mathrm{sd}}}}+\varepsilon_T^{\mathrm{max}}.
    \end{equation}
    If moreover the argument of the square root in \eqref{eq:SHSB-avg-result} is at least $f_{\mathrm{min}}>0$ for all $T$, the error term becomes $\varepsilon_T$ as in (b).
\end{enumerate}
The constant $c$ only depends on $M$, $C$, $K_4$, $a_0$ (and $f_{\mathrm{min}}$ in (b)-(c)), and not on $n$ nor on the number $r$ of splits.\par
\end{proposition}
\begin{proof}
Since $\hat{\rho}_{A,B}=u_1/\sqrt{u_2u_3}$ for $\hat{\mathbf{Z}}_T:=\left(\widehat{\mathrm{Cov}}_t\left(\mathbf{v}^A,\mathbf{v}^B\right),\widehat{\mathrm{Var}}_t\left(\mathbf{v}^A\right),\widehat{\mathrm{Var}}_t\left(\mathbf{v}^B\right)\right)=:(u_1,u_2,u_3)$, we have $\frac{2\hat{\rho}_{A,B}}{1+\hat{\rho}_{A,B}}=F\left(\hat{\mathbf{Z}}_T\right)$ with
    \begin{equation}
        F(u_1,u_2,u_3):=\frac{2u_1}{u_1+\sqrt{u_2u_3}}.
    \end{equation}
    The two sets $A$ and $B$ are disjoint, so $s(A,B)=0$, and
    \begin{equation}
        s(A,A)=\frac{4}{n^2}\sum_{\ell\in A}\sigma_\ell^2=\frac{2\bar{\sigma}^2_A}{n},\qquad s(B,B)=\frac{2\bar{\sigma}^2_B}{n}.
    \end{equation}
    As a consequence the population counterpart of $\hat{\mathbf{Z}}_T$ is
    \begin{equation}
        \mathbf{z}_T=(z_1,z_2,z_3):=\left(\sum_k\bar{w}_k^A\bar{w}_k^B,\ \sum_k\left(\bar{w}_k^A\right)^2+\frac{2\bar{\sigma}^2_A}{n},\ \sum_k\left(\bar{w}_k^B\right)^2+\frac{2\bar{\sigma}^2_B}{n}\right).
    \end{equation}
    Since $|A|=|B|$, $\bar{w}_k=\frac{1}{2}\left(\bar{w}_k^A+\bar{w}_k^B\right)$ and $\frac{1}{2}\left(\bar{\sigma}^2_A+\bar{\sigma}^2_B\right)=\bar{\sigma}^2$, therefore
    \begin{equation}
        z_2+z_3+2z_1=\sum_k\left(\bar{w}_k^A+\bar{w}_k^B\right)^2+\frac{2\left(\bar{\sigma}^2_A+\bar{\sigma}^2_B\right)}{n}=4\left(\sum_k\bar{w}_k^2+\frac{\bar{\sigma}^2}{n}\right).
    \end{equation}
    Using this, we rewrite the denominator of $F(\mathbf{z}_T)$ as
    \begin{align}
        z_1+\sqrt{z_2z_3}&=z_1+\frac{1}{2}\left(z_2+z_3-\left(\sqrt{z_2}-\sqrt{z_3}\right)^2\right)\nonumber\\
        &=\frac{1}{2}\left(2z_1+z_2+z_3\right)-\frac{1}{2}\left(\sqrt{z_2}-\sqrt{z_3}\right)^2\nonumber\\
        &=2\left(\sum_k\bar{w}_k^2+\frac{\bar{\sigma}^2}{n}-\Delta_{AB}^{\mathrm{sd}}\right).
        \label{eq:den_shsb}
    \end{align}
    The numerator can be developed as
    \begin{equation}
        2\sum_k\bar{w}_k^A \bar{w}_k^B=\frac{1}{2}\sum_k\left(\left(\bar{w}_k^A+\bar{w}_k^B\right)^2-\left(\bar{w}_k^A-\bar{w}_k^B\right)^2\right)=2\left(\sum_k\bar{w}_k^2-\Delta_{AB}^{w}\right).
        \label{eq:num_shsb}
    \end{equation}
    Combining \eqref{eq:num_shsb} and \eqref{eq:den_shsb} shows that $F(\mathbf{z}_T)$ is equal to the argument of the square root in \eqref{eq:SHSB-result}. To prove the proposition, we need to check that a clipped version of $F$ satisfies the conditions of Lemma~\ref{lem:delta-method}, and then use the results of Sec.~\ref{sec:large_T_convergence}.
\begin{enumerate}[(1)]
    \item \textbf{Compact set, smoothness and boundedness} As in the proof of Corollary~\ref{cor:correlation-of-trial-averages}, $z_1\leq M^2$ and $z_2,z_3\in\left[a_0,M^2+C\right]$. Moreover, $z_1\geq0$ by \eqref{eq:shsb-cond} and, by the Cauchy-Schwarz inequality, $z_1=\sum_k\bar{w}_k^A\bar{w}_k^B\leq\sqrt{\sum_k\left(\bar{w}_k^A\right)^2\sum_k\left(\bar{w}_k^B\right)^2}\leq\sqrt{z_2z_3}$. Hence $\mathbf{z}_T$ lies in the compact set $\mathcal{K}:=\left[0,M^2\right]\times\left[a_0,M^2+C\right]\times\left[a_0,M^2+C\right]$, and $0\leq F\left(\mathbf{z}_T\right)\leq1$. We take $\delta:=a_0/8$ so that every point $(u_1,u_2,u_3)$ of $\mathcal{K}_\delta$ satisfies $u_1\geq-a_0/8$ and $u_2,u_3\geq 7a_0/8$, and therefore lies in the open set
    \begin{equation}
        \mathcal{O}:=\left\{(u_1,u_2,u_3):u_1>-a_0/4,\ u_2>3a_0/4,\ u_3>3a_0/4\right\}.
    \end{equation}
    On $\mathcal{O}$, $u_1+\sqrt{u_2u_3}>a_0/2>0$, so that $F$ is infinitely differentiable on $\mathcal{O}$. Moreover, $F>-1$ on $\mathcal{O}$: multiplying by the positive denominator, $F(u_1,u_2,u_3)>-1$ is equivalent to $3u_1+\sqrt{u_2u_3}>0$, which holds since $3u_1>-3a_0/4$ and $\sqrt{u_2u_3}>3a_0/4$. To bound $F\left(\hat{\mathbf{Z}}_T\right)$, we define $\tilde{F}:=\mathrm{max}\left(F,-1\right)$, with $\tilde{F}:=0$ if $u_2u_3\leq0$, and $\tilde{F}:=-1$ if $u_2u_3>0$ and $u_1+\sqrt{u_2u_3}=0$. It follows that $\tilde{F}=F$ on $\mathcal{O}$, so that $\tilde{F}$ is infinitely differentiable on the open set $\mathcal{O}$ containing $\mathcal{K}_\delta$ and $\left|\tilde{F}\left(\mathbf{z}_T\right)\right|\leq1$. Moreover, $\left|\tilde{F}\left(\hat{\mathbf{Z}}_T\right)\right|\leq1$ since $\hat{\rho}_{A,B}\in[-1,1]$ and $\frac{2\rho}{1+\rho}\leq1$ for $-1<\rho\leq1$. Hence $\tilde{F}$, $\mathcal{K}$, $\delta$ and $U=1$ satisfy the conditions of Lemma~\ref{lem:delta-method}. Finally, since $\tilde{F}$ and $F$ coincide where they are positive, $\widehat{NC}^{\mathrm{SHSB}}_{A,B}=\sqrt{\mathrm{max}\left(0,\tilde{F}\left(\hat{\mathbf{Z}}_T\right)\right)}$.
    \item \textbf{Proof of (a)} Theorem~\ref{thm:general-limit-theorem} applied to $\tilde{F}$ with $\mathcal{K}$, $\delta$ and $U=1$ gives $\mathbb{E}\left[\left(\tilde{F}\left(\hat{\mathbf{Z}}_T\right)-F\left(\mathbf{z}_T\right)\right)^2\right]\leq c/T$, and Lemma~\ref{lem:thresholded-functions} with $X_T=\tilde{F}\left(\hat{\mathbf{Z}}_T\right)$ and $x_T=F\left(\mathbf{z}_T\right)$ gives (a).
    \item \textbf{Proof of (b)} Let $\tilde{\mathcal{K}}:=\left\{\mathbf{u}\in\mathcal{K}:F(\mathbf{u})\geq f_{\mathrm{min}}\right\}$, which is compact and contains all $\mathbf{z}_T$. Since $\mathcal{K}_{\delta}$ is a compact subset of $\mathcal{O}$, on which $F$ is continuously differentiable, $L:=\mathrm{sup}_{\mathbf{u}\in\mathcal{K}_{\delta}}|\nabla F(\mathbf{u})|<\infty$. We take $\tilde\delta:=\mathrm{min}\left(\delta,\,f_{\mathrm{min}}/(2L)\right)$. For $\mathbf{y}\in\tilde{\mathcal{K}}_{\tilde\delta}$, there exists $\mathbf{x}\in\tilde{\mathcal{K}}\subset\mathcal{K}$ with $|\mathbf{x}-\mathbf{y}|\leq\tilde\delta\leq\delta$. The segment between $\mathbf{x}$ and $\mathbf{y}$ lies in $\mathcal{K}_{\delta}$, and the mean value theorem gives $|F(\mathbf{y})-F(\mathbf{x})|\leq L\tilde\delta\leq f_{\mathrm{min}}/2$, so that $F(\mathbf{y})\geq f_{\mathrm{min}}/2$. Hence, on the open set $\mathcal{O}\cap\left\{F>f_{\mathrm{min}}/4\right\}$, which contains $\tilde{\mathcal{K}}_{\tilde\delta}$, $G:=\sqrt{\mathrm{max}\left(0,\tilde{F}\right)}=\sqrt{F}$ is infinitely differentiable. Moreover, $0\leq G\left(\hat{\mathbf{Z}}_T\right)\leq1$ and $0\leq G\left(\mathbf{z}_T\right)\leq1$ by step (1). Theorem~\ref{thm:general-limit-theorem} applied to $G$ with $\tilde{\mathcal{K}}$, $\tilde\delta$ and $U=1$ gives (b).
    \item \textbf{Proof of (c)} Write $\mathcal{S}=\{(A_1,B_1),...,(A_r,B_r)\}$, and let $\hat{\mathbf{Z}}^{(j)}_T$ and $\mathbf{z}^{(j)}_T$ be the vectors defined at the beginning of the proof for the split $(A_j,B_j)$, so that
    \begin{equation}
        \widehat{NC}^{\mathrm{SHSB}}=g\left(\frac{1}{r}\sum_{j=1}^rF\left(\hat{\mathbf{Z}}^{(j)}_T\right)\right),\qquad g(x):=\sqrt{\mathrm{max}(0,x)},
    \end{equation}
    We proceed in three steps: we first replace $F$ by its clipped version $\tilde{F}$ in every split, then average the expansions of the individual splits, and finally apply $g$.
    \begin{enumerate}[(i)]
        \item \textbf{Clipping} Let $X:=\frac{1}{r}\sum_jF\left(\hat{\mathbf{Z}}^{(j)}_T\right)$ and $\tilde{X}:=\frac{1}{r}\sum_j\tilde{F}\left(\hat{\mathbf{Z}}^{(j)}_T\right)$. Since $F\leq\tilde{F}$ and $g$ is non-decreasing and takes values in $[0,1]$ on $(-\infty,1]$, the difference $D_T:=g(X)-g(\tilde{X})$ satisfies $-1\leq D_T\leq0$, and $D_T=0$ unless $F\left(\hat{\mathbf{Z}}^{(j)}_T\right)<-1$ for some $j$. By step (1), $F>-1$ on $\mathcal{O}\supset\mathcal{K}_\delta$, so this requires $\hat{\mathbf{Z}}^{(j)}_T\notin\mathcal{K}_\delta$, and hence, since $\mathbf{z}^{(j)}_T\in\mathcal{K}$, $\left|\hat{\mathbf{Z}}^{(j)}_T-\mathbf{z}^{(j)}_T\right|>\delta$, for some $j$. We bound the probability of this event uniformly in $r$ and $n$. Each entry of $\hat{\mathbf{Z}}^{(j)}_T-\mathbf{z}^{(j)}_T$ is of the form
        \begin{equation}
            \frac{4}{n^2}\sum_{\ell\in A'}\sum_{\ell'\in B'}\left(\widehat{\mathrm{Cov}}_t\left(\mathbf{v}^{(\ell)},\mathbf{v}^{(\ell')}\right)-\overline{\mathrm{Cov}}\left(\mathbf{v}^{(\ell)},\mathbf{v}^{(\ell')}\right)\right),\qquad A',B'\in\{A_j,B_j\},
        \end{equation}
        and, by (4) of the proof of Theorem~\ref{thm:general-limit-theorem}, each difference in the sum decomposes as $Y_{\ell\ell'}+W_{\ell\ell'}$ with $\mathbb{E}\left[Y_{\ell\ell'}^2\right]\leq c_1'/T$ and $\mathbb{E}\left[|W_{\ell\ell'}|\right]\leq c_2'/T$, where $c_1':=\left(4M\sqrt{C}+\sqrt{C^2+K_4}\right)^2$ and $c_2':=C$. Bounding the sum over $A'\times B'$ by the sum over all pairs $(\ell,\ell')\in[n]^2$, which does not depend on $j$, every entry of every split satisfies
        \begin{align}
            &\mathrm{max}_{j}\left|\left[\hat{\mathbf{Z}}^{(j)}_T-\mathbf{z}^{(j)}_T\right]_q\right|\leq\bar{Y}+\bar{W},\,\mathrm{for}\,q=1,2,3,\\&\bar{Y}:=\frac{4}{n^2}\sum_{\ell,\ell'=1}^n|Y_{\ell\ell'}|,\qquad\bar{W}:=\frac{4}{n^2}\sum_{\ell,\ell'=1}^n|W_{\ell\ell'}|,
        \end{align}
        where Minkowski's inequality gives $\mathbb{E}\left[\bar{Y}^2\right]\leq16c_1'/T$. Additionally, $\mathbb{E}\left[\bar{W}\right]\leq4c_2'/T$. Since $\hat{\mathbf{Z}}^{(j)}_T-\mathbf{z}^{(j)}_T$ has three entries, $D_T\neq 0$ implies $\bar{Y}+\bar{W}>\frac{\delta}{\sqrt{3}}$ (since we need at least one component $q$ with $\left|\left[\hat{\mathbf{Z}}^{(j)}_T-\mathbf{z}^{(j)}_T\right]_q\right|>\delta/\sqrt{3}$).  Furthermore $\bar{Y}+\bar{W}>\delta/\sqrt{3}$ requires $\bar{Y}>\frac{\delta}{2\sqrt3}$ or $\bar{W}>\frac{\delta}{2\sqrt3}$. Markov's inequality then gives \begin{align}
            \mathbb{P}\left(D_T\neq0\right)&\leq\mathbb{P}\left(\bar{Y}>\frac{\delta}{2\sqrt{3}}\right)+\mathbb{P}\left(\bar{W}>\frac{\delta}{2\sqrt{3}}\right)\nonumber\\
            &\leq\frac{\mathbb{E}\left[\bar{Y}^2\right]}{\delta^2/12}+\frac{\mathbb{E}\left[\bar{W}\right]}{\delta/(2\sqrt{3})}\nonumber\\
            &\leq\frac{192c_1'}{\delta^2T}+\frac{8\sqrt{3}c_2'}{\delta T}=:\frac{c_D}{T}.
        \end{align}
        Hence, since $|D_T|\leq1$, $\left|\mathbb{E}\left[D_T\right]\right|\leq\mathbb{E}|D_T|\leq\sup|D_T|\,\mathbb{P}(D_T\neq 0)\leq c_D/T$, and similarly $\mathbb{E}\left[D_T^2\right]\leq c_D/T$.
        \item \textbf{Averaging over splits} By Theorem~\ref{thm:general-limit-theorem} applied to $\tilde{F}$ as in step (2), for every split, $\tilde{F}\left(\hat{\mathbf{Z}}^{(j)}_T\right)=F\left(\mathbf{z}^{(j)}_T\right)+\varepsilon^{(j)}_T$ with $\left|\mathbb{E}\left[\varepsilon^{(j)}_T\right]\right|\leq c/T$ and $\mathbb{E}\left[\left(\varepsilon^{(j)}_T\right)^2\right]\leq c/T$, where $c$ is the same for all splits, since it only depends on $M$, $C$, $K_4$ and $a_0$. Averaging, we have
        \begin{equation}
            \tilde{X}=\bar{x}_T+\bar{\varepsilon}_T,\qquad\bar{x}_T:=\frac{1}{r}\sum_{j=1}^rF\left(\mathbf{z}^{(j)}_T\right),\qquad\bar{\varepsilon}_T:=\frac{1}{r}\sum_{j=1}^r\varepsilon^{(j)}_T,
        \end{equation}
        where $\bar{x}_T$ is the argument of the square root in \eqref{eq:SHSB-avg-result}. By the triangle inequality $\left|\mathbb{E}\left[\bar{\varepsilon}_T\right]\right|\leq\frac{1}{r}\sum_j\left|\mathbb{E}\left[\varepsilon^{(j)}_T\right]\right|\leq c/T$, and by Minkowski's inequality ${\mathbb{E}\left[\bar{\varepsilon}_T^2\right]}\leq\left(\frac{1}{r}\sum_j\sqrt{\mathbb{E}\left[\left(\varepsilon^{(j)}_T\right)^2\right]}\right)^2\leq{c/T}$.
        \item \textbf{Square root} In the general case, since $\mathbb{E}\left[\left(\tilde{X}-\bar{x}_T\right)^2\right]=\mathbb{E}\left[\bar{\varepsilon}_T^2\right]\leq c/T$ by (ii), Lemma~\ref{lem:thresholded-functions} with $X_T=\tilde{X}$ and $x_T=\bar{x}_T$ gives $g(\tilde{X})=g(\bar{x}_T)+\varepsilon^{\mathrm{max}}_T$ with $\left|\mathbb{E}\left[\varepsilon^{\mathrm{max}}_T\right]\right|\leq(c/T)^{1/4}$ and $\mathbb{E}\left[\left(\varepsilon^{\mathrm{max}}_T\right)^2\right]\leq(c/T)^{1/2}$. If $\bar{x}_T\geq f_{\mathrm{min}}$ for all $T$, we apply Lemma~\ref{lem:delta-method} with $d=1$ instead. The expansion of (ii) has the form required by this lemma, with the random part $Y_T:=\bar{\varepsilon}_T-\mathbb{E}\left[\bar{\varepsilon}_T\right]$ and the deterministic part $W_T:=\mathbb{E}\left[\bar{\varepsilon}_T\right]$: we have $\mathbb{E}\left[Y_T\right]=0$, $\mathbb{E}\left[Y_T^2\right]\leq\mathbb{E}\left[\bar{\varepsilon}_T^2\right]\leq c/T$ and $\mathbb{E}\left[|W_T|\right]=\left|\mathbb{E}\left[\bar{\varepsilon}_T\right]\right|\leq c/T$. We take $\mathcal{K}=\left[f_{\mathrm{min}},1\right]$, which contains all $\bar{x}_T$ since $0\leq F\leq1$ at the population counterparts (step (1)), and $\delta=f_{\mathrm{min}}/2$. Then $g$ is infinitely differentiable on the open set $\left(f_{\mathrm{min}}/4,\infty\right)\supset\mathcal{K}_\delta$, and $0\leq g(\tilde{X}),g(\bar{x}_T)\leq1$ since $\tilde{X}\leq1$. Lemma~\ref{lem:delta-method} then gives $g(\tilde{X})=g(\bar{x}_T)+\varepsilon_T$.
    \end{enumerate}
    Finally, $\widehat{NC}^{\mathrm{SHSB}}=g(X)=g(\bar{x}_T)+\varepsilon^{\mathrm{max}}_T+D_T$ (respectively $g(\bar{x}_T)+\varepsilon_T+D_T$). By (i), $\left|\mathbb{E}\left[D_T\right]\right|\leq c_D/T$ and $\mathbb{E}\left[D_T^2\right]\leq c_D/T$. By the triangle inequality for the bias and Minkowski’s inequality for the root mean squared error, adding $D_T$ to $\varepsilon_T$ gives an error term of type $\varepsilon_T$ with constant $(\sqrt{c}+\sqrt{c_D})^2$. Adding $D_T$ to $\varepsilon^{\mathrm{max}}_T$ gives an error term of type $\varepsilon^{\mathrm{max}}_T$ with constant $\left(c^{1/4}+\mathrm{max}\left(c_D,\sqrt{c_D}\right)\right)^4$.
 
\end{enumerate}
 
\end{proof}
 
\begin{lemma}[average over all splits]\label{lem:split-average} Let $\mathcal{S}$ be the collection of all possible splits. Then
\begin{equation}\label{eq:split-average}
    \frac{1}{|\mathcal{S}|}\sum_{(A,B)\in\mathcal{S}}\Delta_{AB}^w=\frac{1}{n-1}\sum_k\widehat{\mathrm{Var}}_\ell\left[w_k\right].
\end{equation}
Hence, the average over all splits of the numerator of \eqref{eq:SHSB-result}, $\sum_k\bar{w}_k^2-\Delta_{AB}^w$, is exactly the numerator $\sum_k\bar{w}_k^2-\frac{1}{n-1}\sum_k\widehat{\mathrm{Var}}_\ell\left[w_k\right]$ of $\widehat{NC}^{\mathrm{ATV}}$.
\end{lemma}
\begin{proof}
For a split $(A,B)$ we define $\epsilon_\ell^{A}:=1$ if $\ell\in A$ and $-1$ otherwise. We develop
\begin{align}
    \frac{1}{|\mathcal{S}|}\sum_{(A,B)\in\mathcal{S}}\Delta_{AB}^w&=\frac{1}{|\mathcal{S}|}\sum_{(A,B)\in\mathcal{S}}\frac{1}{4}\sum_k\left(\bar{w}_k^A-\bar{w}_k^B\right)^2\nonumber\\
    &=\frac{1}{4}\frac{1}{|\mathcal{S}|}\sum_{(A,B)\in\mathcal{S}}\sum_k\left(\frac{2}{n}\sum_\ell\epsilon_\ell^{A}w_k^{(\ell)}\right)^2\nonumber\\
    &=\frac{1}{n^2}\frac{1}{|\mathcal{S}|}\sum_{(A,B)\in\mathcal{S}}\sum_k\left(\sum_\ell\left(w_k^{(\ell)}\right)^2+\sum_{\ell\neq\ell'}\epsilon_\ell^{A}\epsilon_{\ell'}^{A}w_k^{(\ell)}w_k^{(\ell')}\right)\nonumber\\
    &=\frac{1}{n^2}\sum_k\left(\sum_\ell\left(w_k^{(\ell)}\right)^2+\sum_{\ell\neq\ell'}w_k^{(\ell)}w_k^{(\ell')}\frac{1}{|\mathcal{S}|}\sum_{(A,B)\in\mathcal{S}}\epsilon_\ell^{A}\epsilon_{\ell'}^{A}\right).\label{eq:delta_AB_dev}
\end{align}
By definition, $\epsilon_\ell^{A}\epsilon_{\ell'}^{A}$ is $1$ if $\ell$ and $\ell'$ are either both in $A$ or both in $B$; it is $-1$ otherwise. Using this we write
\begin{align}
    \frac{1}{|\mathcal{S}|}\sum_{(A,B)\in\mathcal{S}}\epsilon_\ell^{A}\epsilon_{\ell'}^{A}&=\frac{1}{|\mathcal{S}|}\sum_{(A,B)\in\mathcal{S}}\mathbbm{1}_{\{\ell,\ell'\in A\,\mathrm{or}\,\ell,\ell'\in B\}}-\left(1-\frac{1}{|\mathcal{S}|}\sum_{(A,B)\in\mathcal{S}}\mathbbm{1}_{\{\ell,\ell'\in A\,\mathrm{or}\,\ell,\ell'\in B\}}\right)\nonumber\\
    &=\frac{2}{|\mathcal{S}|}\sum_{(A,B)\in\mathcal{S}}\mathbbm{1}_{\{\ell,\ell'\in A\,\mathrm{or}\,\ell,\ell'\in B\}}-1=4\cdot\frac{\binom{n-2}{n/2-2}}{\binom{n}{n/2}}-1=4\cdot\frac{n/2(n/2-1)}{n(n-1)}-1\nonumber\\
    &=-\frac{1}{n-1}.\label{eq:epsilon_prod}
\end{align}
Injecting \eqref{eq:epsilon_prod} in \eqref{eq:delta_AB_dev},
\begin{align}
    \frac{1}{|\mathcal{S}|}\sum_{(A,B)\in\mathcal{S}}\Delta_{AB}^w&=\frac{1}{n^2}\sum_k\left(\sum_\ell\left(w_k^{(\ell)}\right)^2-\frac{1}{n-1}\sum_{\ell\neq\ell'}w_k^{(\ell)}w_k^{(\ell')}\right)\nonumber\\
    &=\frac{1}{n^2}\sum_k\left(\frac{n}{n-1}\sum_\ell\left(w_k^{(\ell)}\right)^2-\frac{1}{n-1}\left(\sum_{\ell}w_k^{(\ell)}\right)^2\right)\nonumber\\
    &=\frac{1}{n-1}\sum_k\left(\frac{1}{n}\sum_{\ell}\left(w_k^{(\ell)}\right)^2-\left(\frac{1}{n}\sum_{\ell}w_k^{(\ell)}\right)^2\right)\nonumber\\
    &=\frac{1}{n-1}\sum_k\widehat{\mathrm{Var}}_\ell\left[w_k\right].
\end{align}
\end{proof}
 
\begin{remark}[comparison with ATV] For $\mathcal{S}$ being the collection of all splits, using Lemma~\ref{lem:split-average},
\begin{align}
    \sqrt{\frac{1}{r}\sum_{(A,B)\in\mathcal{S}}\frac{\sum_k\bar{w}_k^2-\Delta_{AB}^{w}}{\sum_k\bar{w}_k^2+\bar{\sigma}^2/n-\Delta_{AB}^{\mathrm{sd}}}}&\geq\sqrt{\frac{1}{r}\sum_{(A,B)\in\mathcal{S}}\frac{\sum_k\bar{w}_k^2-\Delta_{AB}^{w}}{\sum_k\bar{w}_k^2+\bar{\sigma}^2/n}}\nonumber\\
    &=\sqrt{\frac{\sum_k\bar{w}_k^2-\frac{1}{n-1}\sum_k\widehat{\mathrm{Var}}_\ell\left[w_k\right]}{\sum_k\bar{w}_k^2+\bar{\sigma}^2/n}}.
\end{align}
This implies that, up to their error terms, $\widehat{NC}^{\mathrm{SHSB}}\geq\widehat{NC}^{\mathrm{ATV}}$. The equality holds if and only if $\Delta_{AB}^{\mathrm{sd}}=0$ for each split $(A,B)$. 
\end{remark}

\subsection{Single-trial noise ceiling}\label{sec:single_trial_noise_ceiling}
In this section, we derive the closed-form expressions for the true noise ceiling for single-trial responses $NC^{\mathrm{single}}_\ell$ and its estimators $\widehat{NC}^{\mathrm{LOOC}}_\ell$ and $\widehat{NC}^{\mathrm{TTMC}}_\ell$. We fix a trial $\ell$ and assume $n\geq2$. We write
\begin{equation}
    \mathbf{v}^{(-\ell)}:=\mathbf{v}^{[n]\setminus\{\ell\}},\qquad\mathbf{v}^{*(-\ell)}:=\mathbf{v}^{*[n]\setminus\{\ell\}},\qquad\bar{w}_k^{(-\ell)}:=\bar{w}_k^{[n]\setminus\{\ell\}},\qquad\bar{\sigma}^2_{(-\ell)}:=\frac{1}{n-1}\sum_{\ell'\neq\ell}\sigma_{\ell'}^2
\end{equation}
for the average of the other trials, its noise-free counterpart, its weights and its noise level.
 
\subsubsection{True noise ceiling ${NC}^{\mathrm{single}}$}\label{sec:true_noise_ceiling_single}
The true noise ceiling for single-trial responses is the correlation between the noise-free signal and the recorded signal of the same trial, $NC^{\mathrm{single}}_\ell:=\widehat{\mathrm{Corr}}_t\left(\mathbf{v}^{*(\ell)},\mathbf{v}^{(\ell)}\right)$.
 
\begin{proposition}[true noise ceiling, single-trial responses]\label{prop:ncsingle} Assume that there exists a constant $a_0>0$ such that $\widehat{\mathrm{Var}}_t\left(\mathbf{v}^{*(\ell)}\right)\geq a_0$ for all $T$. Under Assumptions~\ref{ass:independent-trials}--\ref{ass:summable-fourth-cumulants},
\begin{equation}\label{eq:nc_single_app}
    NC^{\mathrm{single}}_\ell=\sqrt{\frac{\widehat{\mathrm{Var}}_t\left(\mathbf{v}^{*(\ell)}\right)}{\widehat{\mathrm{Var}}_t\left(\mathbf{v}^{*(\ell)}\right)+\sigma_\ell^2}}+\varepsilon_T=\sqrt{\frac{\sum_k\left(w_k^{(\ell)}\right)^2}{\sum_k\left(w_k^{(\ell)}\right)^2+\sigma_\ell^2}}+\varepsilon_T,
\end{equation}
where the constant $c$ of the error term only depends on $M$, $C$, $K_4$ and $a_0$. The second expression holds under the representation of Definition~\ref{def:latent-signals}. In contrast to $NC^{\mathrm{mean}}$, the leading term does not depend on $n$: it is determined by the signal and the noise of trial $\ell$ alone.
\end{proposition}
\begin{proof}
We apply \eqref{eq:corr-free} of Corollary~\ref{cor:correlation-of-trial-averages} with $A=B=\{\ell\}$, for which $s(\{\ell\},\{\ell\})=\sigma_\ell^2$, and simplify as in the proof of Proposition~\ref{prop:true-noise-ceiling-trial-averaged}. The second expression follows from Lemma~\ref{lem:latent-form-of-the} with $A=\{\ell\}$.
\end{proof}
 
\subsubsection{Leave-one-out correlation $\widehat{NC}^{\mathrm{LOOC}}$}\label{sec:LOOC}
The LOOC estimator correlates trial $\ell$ with the average of the other trials, $\widehat{NC}^{\mathrm{LOOC}}_\ell:=\widehat{\mathrm{Corr}}_t\left(\mathbf{v}^{(\ell)},\mathbf{v}^{(-\ell)}\right)$.
 
\begin{proposition}[LOOC]\label{prop:looc} Assume that there exists a constant $a_0>0$ such that $\widehat{\mathrm{Var}}_t\left(\mathbf{v}^{*(\ell)}\right)\geq a_0$ and $\widehat{\mathrm{Var}}_t\left(\mathbf{v}^{*(-\ell)}\right)\geq a_0$ for all $T$. Under Assumptions~\ref{ass:independent-trials}--\ref{ass:summable-fourth-cumulants},
\begin{equation}\label{eq:LOOC}
    \widehat{NC}^{\mathrm{LOOC}}_\ell=\frac{\sum_kw_k^{(\ell)}\bar{w}_k^{(-\ell)}}{\sqrt{\left(\sum_k\left(w_k^{(\ell)}\right)^2+\sigma_\ell^2\right)\left(\sum_k\left(\bar{w}_k^{(-\ell)}\right)^2+\frac{\bar{\sigma}^2_{(-\ell)}}{n-1}\right)}}+\varepsilon_T
\end{equation}
where the constant $c$ of the error term only depends on $M$, $C$, $K_4$ and $a_0$.
\end{proposition}
\begin{proof}
We apply \eqref{eq:corr-rec} of Corollary~\ref{cor:correlation-of-trial-averages} with $A=\{\ell\}$ and $B=[n]\setminus\{\ell\}$. The two sets are disjoint, so $s(A,B)=0$, while $s(A,A)=\sigma_\ell^2$ and
\begin{equation}
    s(B,B)=\frac{1}{(n-1)^2}\sum_{\ell'\neq\ell}\sigma_{\ell'}^2=\frac{\bar{\sigma}^2_{(-\ell)}}{n-1}.
\end{equation}
Lemma~\ref{lem:latent-form-of-the} gives $\widehat{\mathrm{Cov}}_t\left(\mathbf{v}^{*(\ell)},\mathbf{v}^{*(-\ell)}\right)=\sum_kw_k^{(\ell)}\bar{w}_k^{(-\ell)}$, $\widehat{\mathrm{Var}}_t\left(\mathbf{v}^{*(\ell)}\right)=\sum_k\left(w_k^{(\ell)}\right)^2$ and $\widehat{\mathrm{Var}}_t\left(\mathbf{v}^{*(-\ell)}\right)=\sum_k\left(\bar{w}_k^{(-\ell)}\right)^2$.
\end{proof}
\begin{remark}
In the limit of large $n$, $\widehat{NC}^{\mathrm{LOOC}}_\ell$ approaches
\begin{equation}
    NC_\ell^{\mathrm{single}}\cdot\cos_k\left(w^{(\ell)},\overline{w}^{(-\ell)}\right)\leq NC_\ell^{\mathrm{single}},
\end{equation}
up to an error term $\varepsilon_T$, and with
\begin{equation}
    \cos_k\left(x,y\right):=\frac{\sum_kx_ky_k}{\sqrt{\sum_kx_k^2\sum_ky_k^2}}
\end{equation}
the cosine similarity across $k$. The equality holds if and only if $w^{(\ell)}=\lambda\,\overline{w}^{(-\ell)}$ for some $\lambda>0$, which is in general satisfied only in the homogeneous regime.
\end{remark}
\subsubsection{Trial-to-mean correlation $\widehat{NC}^{\mathrm{TTMC}}$}\label{sec:TTMC}
The TTMC estimator correlates trial $\ell$ with the average of all trials, including trial $\ell$ itself, $\widehat{NC}^{\mathrm{TTMC}}_\ell:=\widehat{\mathrm{Corr}}_t\left(\mathbf{v}^{(\ell)},\bar{\mathbf{v}}\right)$.
 
\begin{proposition}[TTMC]\label{prop:ttmc} Assume that there exists a constant $a_0>0$ such that $\widehat{\mathrm{Var}}_t\left(\mathbf{v}^{*(\ell)}\right)\geq a_0$ and $\widehat{\mathrm{Var}}_t\left(\bar{\mathbf{v}}^{*}\right)\geq a_0$ for all $T$. Under Assumptions~\ref{ass:independent-trials}--\ref{ass:summable-fourth-cumulants},
\begin{equation}\label{eq:TTMC}
    \widehat{NC}^{\mathrm{TTMC}}_\ell=\frac{\sum_kw_k^{(\ell)}\bar{w}_k+\frac{\sigma_\ell^2}{n}}{\sqrt{\left(\sum_k\left(w_k^{(\ell)}\right)^2+\sigma_\ell^2\right)\left(\sum_k\bar{w}_k^2+\frac{\bar{\sigma}^2}{n}\right)}}+\varepsilon_T
\end{equation}
where the constant $c$ of the error term only depends on $M$, $C$, $K_4$ and $a_0$.\par
\end{proposition}
\begin{proof}
We apply \eqref{eq:corr-rec} of Corollary~\ref{cor:correlation-of-trial-averages} with $A=\{\ell\}$ and $B=[n]$. Now $A\cap B=\{\ell\}$, so that
\begin{equation}
    s(A,B)=\frac{\sigma_\ell^2}{n},\qquad s(A,A)=\sigma_\ell^2,\qquad s(B,B)=\frac{1}{n^2}\sum_{\ell'=1}^n\sigma_{\ell'}^2=\frac{\bar{\sigma}^2}{n}.
\end{equation}
Lemma~\ref{lem:latent-form-of-the} gives $\widehat{\mathrm{Cov}}_t\left(\mathbf{v}^{*(\ell)},\bar{\mathbf{v}}^{*}\right)=\sum_kw_k^{(\ell)}\bar{w}_k$.
\end{proof}
\begin{remark}
In the limit of large $n$, $\widehat{NC}^{\mathrm{TTMC}}_\ell$ approaches
\begin{equation}
    NC_\ell^{\mathrm{single}}\cdot\cos_k\left(w^{(\ell)},\overline{w}\right)\leq NC_\ell^{\mathrm{single}},
\end{equation}
up to an error term $\varepsilon_T$.
Hence, similarly to $\widehat{NC}^{\mathrm{LOOC}}_\ell$, $\widehat{NC}^{\mathrm{TTMC}}_\ell$ only converges to the true noise ceiling in the homogeneous regime.
\end{remark}

\begin{remark}
In the heterogeneous regime, $\widehat{NC}^{\mathrm{TTMC}}_\ell$ is obviously not an upper bound on the noise ceiling as it converges to a limit that is lower than $NC_\ell^{\mathrm{single}}$. Moreover, even in the homogeneous case,
\begin{equation}
    \widehat{NC}^{\mathrm{TTMC}}_\ell=NC_\ell^{\mathrm{single}}\cdot\underbrace{\frac{1+\frac{\sigma_\ell^2}{n\sum_kw_k^2}}{\sqrt{1+\frac{\overline{\sigma}^2}{n\sum_kw_k^2}}}}_{\gamma}+\varepsilon_T,
\end{equation}
where $\gamma$ can become smaller than $1$ if $2\sigma_\ell^2<\overline{\sigma}^2$.
\end{remark}
\subsubsection{Leave-one-out prediction $\widehat{NC}^{\mathrm{LOOP}}_\ell$}
\label{sec:LOOP}
Since LOOP involves a fitted regression, it does not admit a general closed form in our framework. We describe here its asymptotic behavior qualitatively.
In contrast to the other estimators, LOOP predicts a unit from all units of the other trials. In this section we therefore keep the unit index $i=1,...,m$. The assumptions of Sec.~\ref{sec:large_T_convergence} are taken to hold for every unit, we allow the noise of different units of the same trial to be correlated. Similarly to Sec.~\ref{sec:ATV}, we assume that each recording is centered over time.\par
For a target unit $i$ and trial $\ell$, the $p:=m(n-1)$ predictors are the recordings $\mathbf{v}_j^{(\ell')}$ of all units $j\in[m]$ of all other trials $\ell'\neq\ell$. A linear model with intercept is fitted by ordinary least squares (OLS) or ridge regression on a set of training time points, and $\widehat{NC}^{\mathrm{LOOP}}_{i,\ell}$ is the correlation between its prediction and $\mathbf{v}_i^{(\ell)}$ on a test set of time points. Since the noise of different units of the same trial may be correlated, we extend Assumption~\ref{ass:summable-fourth-cumulants} to pairs of units: for all units $j,k$ and trials $\ell$, $\sup_{t}\sum_{s=1}^\infty\big|\mathrm{Cov}\big(\eta^{(\ell)}_{j,t}\eta^{(\ell)}_{k,t},\,\eta^{(\ell)}_{j,s}\eta^{(\ell)}_{k,s}\big)\big|<K_4$.

\paragraph{Large number of time points}
Let $T$ and $T'$ be the numbers of training and test time points, and
$X\in\mathbb{R}^{T\times p}$, $X'\in\mathbb{R}^{T'\times p}$ the corresponding
predictor matrices. We write $\chi^{(-\ell)}\subset\mathbb{R}^{T'}$ the space spanned by the columns of $X'$ and $P^{(-\ell)}$ for the orthogonal projection onto $\chi^{(-\ell)}$. We consider the limit $T,T'\to\infty$ with $p/T\to0$ and $p/T'\to0$. Assuming the training and test points are chosen at random, $\sigma^2_{i,\ell}$ and $\widehat{\mathrm{Var}}_t(\mathbf{v}_i^{*(\ell)})$ computed over $T$ or $T'$ are equivalent in this limit. The prediction $\hat{\mathbf{v}}_i$ is therefore given by the orthogonal projection $P^{(-\ell)}\mathbf{v}_i^{(\ell)}$ of the target onto $\chi^{(-\ell)}$. Since the noise $\boldsymbol{\eta}_i^{(\ell)}$ of the target is independent of the predictors, the projection of the target is asymptotically equal to $P^{(-\ell)}\mathbf{v}_i^{*(\ell)}$. The estimator $\widehat{NC}^{\mathrm{LOOP}}_{i,\ell}$ is then given by
\begin{equation}\label{eq:LOOP-limit}
    \widehat{\mathrm{Corr}}_t\left(\mathbf{v}_i^{\left(\ell\right)},\hat{\mathbf{v}}_i^{\left(\ell\right)}\right)=\sqrt\frac{{\widehat{\mathrm{Var}}_t\left(P^{(-\ell)}\mathbf{v}_i^{*(\ell)}\right)}}{{\widehat{\mathrm{Var}}_t\left(\mathbf{v}_i^{*(\ell)}\right)+\sigma_{i,\ell}^2}}=\frac{\left|P^{(-\ell)}\mathbf{v}_i^{*(\ell)}\right|}{\left|\mathbf{v}_i^{*(\ell)}\right|}\cdot NC^{\mathrm{single}}_{i,\ell}=\cos\left(\alpha^{(\ell)}\right)\cdot NC^{\mathrm{single}}_{i,\ell},
\end{equation}
where $|.|$ is the Euclidean norm, $\alpha^{(\ell)}$ is the angle between $\mathbf{v}^{*(\ell)}_i$ and $\chi^{(-\ell)}$, and the second equality holds up to the error term of $NC^{\mathrm{single}}_{i,\ell}$. $\widehat{NC}^{\mathrm{LOOP}}_{i,\ell}$ is therefore a lower bound on the single-trial noise ceiling, which is reached if and only if the noise-free target lies in the span of the predictors. Two effects make $\cos\left(\alpha^{(\ell)}\right)<1$. First, the predictors are noisy, therefore $\chi^{(-\ell)}$ is spanned by noisy versions of the reproducible responses rather than by the reproducible responses themselves. Second, in the heterogeneous regime $\mathbf{v}_i^{*(\ell)}$ is likely not contained in the span of the reproducible responses. In particular if $K>p$ the latent signals $\boldsymbol{\phi}_1,...,\boldsymbol{\phi}_K$ cannot be retrieved from $\{\mathbf{v}^{*\ell'}_i\}_{\ell'\neq\ell,\,i=1,...,m}$ and it might not be possible to obtain $\mathbf{v}^{*(\ell)}_i=\sum_kw_{ik}^{(\ell)}\boldsymbol{\phi}_k$ from the reproducible signals. In this limit, $\widehat{NC}^{\mathrm{LOOP}}_{i,\ell}$ is at least as large as $\widehat{NC}^{\mathrm{LOOC}}_{i,\ell}$.

\paragraph{Large number of trials} As $n$ grows, with $T/p\rightarrow\infty$, both effects can decrease: the predictors contain more and more noisy copies of the same latent signals, making it possible to average out their noise, and the reproducible responses of the other trials span a growing subspace. $\alpha^{(\ell)}$ approaches the angle between $\mathbf{v}_i^{*(\ell)}$ and the span of the reproducible responses $\{\mathbf{v}^{*\ell'}_i\}_{\ell'\neq\ell,\,i=1,...,m}$, and $\widehat{NC}^{\mathrm{LOOP}}_{i,\ell}$ recovers the true noise ceiling whenever $\mathbf{v}_i^{*(\ell)}$ lies in this span. In contrast to $\widehat{NC}^{\mathrm{LOOC}}_{i,\ell}$ and $\widehat{NC}^{\mathrm{TTMC}}_{i,\ell}$, the convergence of $\widehat{NC}^{\mathrm{LOOP}}_{i,\ell}$ to ${NC}^{\mathrm{single}}_{i,\ell}$ is not limited to the homogeneous regime.

\paragraph{Finite number of time points} For finite $T$, the generalization error reduces the correlation on the test set. This generalization error is larger when units are very noisy and it grows with the number $p=m(n-1)$ of predictors. For an increasing number of trials $n$ at fixed $T$, $\widehat{NC}^{\mathrm{LOOP}}_{i,\ell}$ contains therefore both an increasing term due to the decay of $\alpha^{(\ell)}$ and a decreasing generalization term. When $p/T$ is not small, it can fall below $\widehat{NC}^{\mathrm{LOOC}}_{i,\ell}$, which has no fitted parameters. Ridge regression reduces the generalization error but replaces it by a bias.

\paragraph{Extrapolation to $n\rightarrow\infty$} These results explain why the dependence of LOOP on $n$ has no simple universal form. The limit term depends, through the angle in \eqref{eq:LOOP-limit}, on how the latent signals of trial $\ell$ are expressed by the other trials. In addition, at fixed $T$, the generalization error grows with $n$. Fitting $\widehat{NC}^{\infty}_{i,\ell}\left(1-e^{-n/\tau_0}\right)$ assumes a very fast saturation and ignores the generalization term, both of which lead to an underestimation of the noise ceiling.

\section{Synthetic data parameters}\label{app:synthetic_param}
Table~\ref{tab:sim_params} gives all parameters used to generate the synthetic datasets used in Fig.~\ref{fig:results}. The parameters are defined in Sec.~\ref{sec:experimental_results}.

\begin{table}[h!]
\centering
\caption{Simulation parameters for each dataset.}
\label{tab:sim_params}
\begin{tabular}{l r r c c c c c l}
\toprule
Dataset & $T$ & $n$ & $\rho_{\text{shared}}$ & $\sigma_0$ & $s$ & $\sigma_w$ & $\Delta$ & Signal \\
\midrule
THINGS fMRI1 & 100  & 12 & 0.30 & 4.00 & 1.0 & 1.00 & 0 & Gaussian \\
THINGS EEG2  & 200  & 80 & 0.15 & 2.75 & 0.9 & 1.00 & 0 & Gaussian \\
Sherlock V1  & 1976 & 16 & 0.60 & 0.60 & 0.6 & 2.00 & 5 & BOLD-like \\
Sherlock IT  & 1976 & 16 & 0.60 & 1.90 & 1.2 & 0.85 & 5 & BOLD-like \\
\bottomrule
\end{tabular}

\vspace{2pt}
{\footnotesize $\Delta = 0$: homogeneous regime; $\Delta > 0$: heterogeneous regime.}
\end{table}

\section{Empirical finite-$T$ convergence}\label{app:finite-T-sherlock}
\begin{figure}[h!]
    \centering
    \includegraphics[width=1\linewidth]{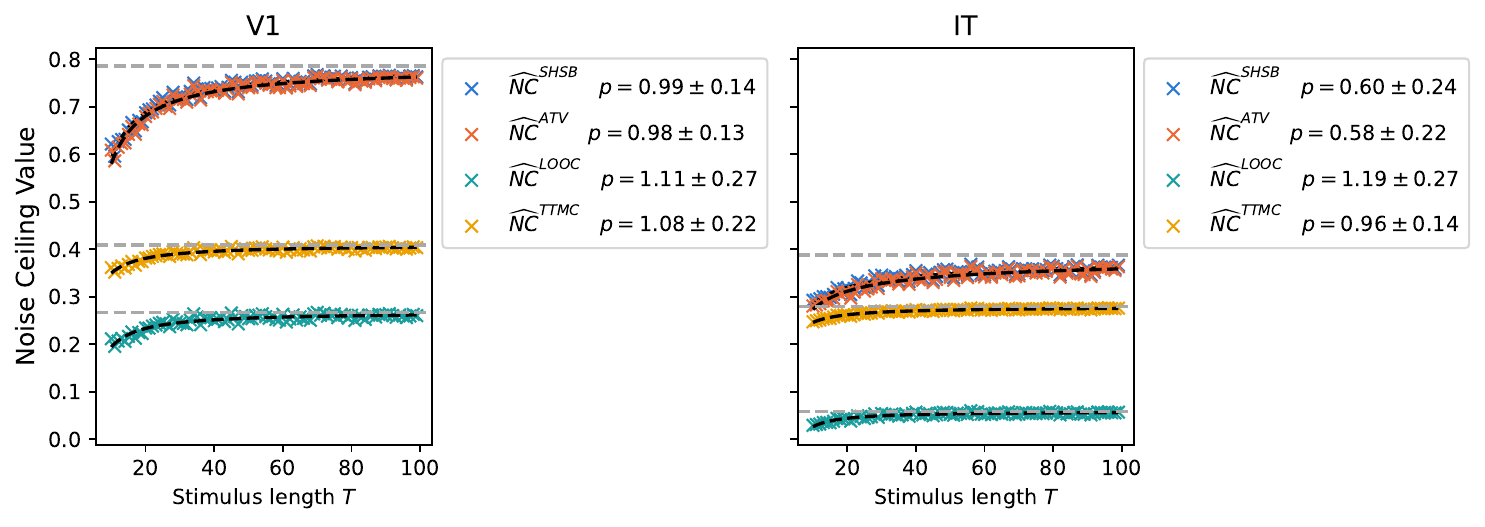}
    \caption{\textbf{Convergence of the estimators with the stimulus length $T$ on the Sherlock dataset.} Each cross is the average value of an estimator computed on 100 time windows of length $T$ and averaged over voxels. All 16 subjects are used at every $T$, so that only the stimulus length $T$ varies. $\widehat{NC}^{\mathrm{LOOC}}$ and $\widehat{NC}^{\mathrm{TTMC}}$ are averaged over every held-out subject. For each estimator, the mean curve is fitted using $\widehat{NC}(T)=a+b\,T^{-p}$ (black dashed lines), where $a$ is the asymptote reached at large $T$ (gray dashed lines) and $p$ the rate of convergence. The legend gives $p$ with a 95\% confidence interval ($\pm1.96$ standard errors, from the covariance of the least-squares fit).
}
    \label{fig:finit-T-shelock}
\end{figure}

In this section, we evaluate empirically the convergence rates of the noise ceiling estimators as the stimulus length $T$ grows on the Sherlock dataset. Fig.~\ref{fig:finit-T-shelock} shows SHSB, ATV, LOOC and TTMC as a function of the stimulus length $T$. If an estimator is the sum of a leading term independent of $T$ and an error whose expectation is of order $1/T$, fitting $a+b\,T^{-p}$ to its mean should give $p\approx1$ once $T$ is large enough. This is what we observe for LOOC and TTMC on both V1 and IT, already for $T<100$. SHSB and ATV behave differently: $p$ is close to $1$ in V1 but lower in IT. This is the transient regime described in Sec.~\ref{sec:results_th_mean}. For units with a weak signal, the threshold of the square root first produces a positive bias of order $T^{-1/4}$, and the $1/T$ rate is reached only once the finite-$T$ fluctuations are small compared to the signal. In V1, where the signal is strong, most voxels are already in the $1/T$ regime over the range of $T$ shown. In IT, where the signal is weaker, the average over voxels mixes voxels with a negative bias of order $1/T$ and voxels with a positive bias of order $T^{-1/4}$. The result is a negative bias that decays more slowly than $1/T$.

 \section{Noise ceiling computation checklist}\label{sec:checklist}
We list here common pitfalls when computing noise ceilings, and how to avoid them.
\paragraph{SHSB and ATV}
\begin{itemize}
    \item \textbf{Square root:} ATV and SHSB both contain a square root. This square root is necessary when the brain score is computed as a Pearson correlation but it is often omitted~\citep{vanbreeHowMuchVariance2025}. The square root should not be applied when the brain score is computed as an $R^2$.
\end{itemize}
\paragraph{ATV}
\begin{itemize}
    \item \textbf{Centering:} The recordings must be centered over time for each trial before the variance across trials is computed.
    \item \textbf{Normalization:} The variance across trials must be computed with the unbiased normalization $1/(n-1)$. With $1/n$, the estimator is biased upwards.
\end{itemize}

\paragraph{SHSB}
\begin{itemize}
    \item \textbf{Averaging over splits:} For SHSB to be a good estimation of the noise ceiling, it needs to be averaged over many splits. To avoid introducing additional biases, this averaging should be performed after the $\frac{2\rho}{1+\rho}$ correction but before the thresholding and square root.
    \item \textbf{Odd number of trials:} For $n$ odd, ATV should be preferred to SHSB.
\end{itemize}

\paragraph{TTMC}
\begin{itemize}
    \item \textbf{Incorrect upper bound:} When trials correspond to different subjects, TTMC should not be used, and should especially not be considered as an upper bound on the noise ceiling.
\end{itemize}

\paragraph{LOOP}
\begin{itemize}
    \item \textbf{Independence of training and test sets:} With autocorrelated recordings, randomly selected training and test time points are correlated. The test set should instead consist of blocks of consecutive time points, separated from the training set by unused gaps longer than the autocorrelation of the signal and noise. The same applies when fitting encoding models.
    \item \textbf{Preprocessing:} Some preprocessing steps, such as high-pass filters, spread information across time points, and can therefore introduce correlations between training and test sets. They should be applied to each set separately.
    \item \textbf{Selecting among estimators:} When taking, for each unit, the highest of LOOC, LOOP and its variants, the selection must be made on a validation set distinct from the test set, to avoid biasing the estimate upwards.
\end{itemize}

\paragraph{Reporting brain scores}
\begin{itemize}
    \item \textbf{Lower bounds:} When only a lower bound on the noise ceiling is available, the normalized brain score should be reported as an interval, from the unnormalized score to the score normalized by the lower bound, clipped at $1$.
\end{itemize}

\end{document}